\documentclass{article}

\PassOptionsToPackage{numbers, compress}{natbib}

\usepackage[preprint]{neurips_2023}

\usepackage[utf8]{inputenc} 
\usepackage[T1]{fontenc}    
\usepackage[hidelinks,colorlinks=true]{hyperref}       
\usepackage{url}            
\usepackage{booktabs}       
\usepackage{amsfonts}       
\usepackage{nicefrac}       
\usepackage{microtype}      
\usepackage{xcolor}         

\usepackage{graphicx}
\usepackage{algorithm}      
\usepackage{algpseudocode} 

\usepackage{amsmath}
\newcommand\norm[1]{\left\lVert#1\right\rVert}

\usepackage{hyperref}
\usepackage{etoc}
\usepackage{chngcntr}
\usepackage{moresize}
\usepackage{enumitem}
\usepackage{caption}
\usepackage{subcaption}

\def\enddisplaymath{\]\@ignoretrue}

\usepackage{titlesec}
\titlespacing*{\section}{0pt}{0.1\baselineskip}{0.2\baselineskip}

\usepackage{arydshln}
\addcontentsline{toc}{section}{Appendices}

\title{ 
RegBN: Batch Normalization of Multimodal Data with Regularization
}

\author{%
    Morteza~Ghahremani$^{1,2}$ \quad Christian~Wachinger$^{1,2}$\\
    $^{1}$Lab for AI in Medical Imaging (AI-Med), Department of Radiology, \\
    Technical University of Munich (TUM), Germany \\ 
    $^{2}$Munich Center for Machine Learning (MCML), Germany\\
    \texttt{\{morteza.ghahremani, christian.wachinger\}@tum.de} \\
}

\begin{document}

\maketitle
\begin{abstract}
Recent years have witnessed a surge of interest in integrating high-dimensional data captured by multisource sensors, driven by the impressive success of neural networks in integrating multimodal data. 
However, the integration of heterogeneous multimodal data poses a significant challenge, as confounding effects and dependencies among such heterogeneous data sources introduce unwanted variability and bias, leading to suboptimal performance of multimodal models. 
Therefore, it becomes crucial to normalize the low- or high-level features extracted from data modalities before their fusion takes place. 
This paper introduces RegBN, a novel approach for multimodal Batch Normalization with REGularization. 
RegBN uses the Frobenius norm as a regularizer term to address the side effects of confounders and underlying dependencies among different data sources. 
The proposed method generalizes well across multiple modalities and eliminates the need for learnable parameters, simplifying training and inference. 
We validate the effectiveness of RegBN on eight databases from five research areas, encompassing diverse modalities such as language, audio, image, video, depth, tabular, and 3D MRI. 
The proposed method demonstrates broad applicability across different architectures such as multilayer perceptrons, convolutional neural networks, and vision transformers, enabling effective normalization of both low- and high-level features in multimodal neural networks.
RegBN is available at \url{https://mogvision.github.io/RegBN}.
\end{abstract}

\addtocontents{toc}{\protect\setcounter{tocdepth}{0}}

\section{Introduction}
Multimodal models, which adeptly fuse information from a diverse range of sources, have yielded promising results and found many applications such as language and vision~\cite{lin2014microsoft,NEURIPS2019_c74d97b0,10006384,huang2022mack,sun2022long}, multimedia~\cite{alayrac2020self,nagrani2021attention,duquenne2021multimodal}, affective computing~\cite{zadeh2016mosi,poria2017review,zadeh2018multimodal,rouast2019deep,verma2023multimodal}, 
robotics~\cite{lee2019making,9756910,lee2020multimodal},
human-computer interaction~\cite{preece2015interaction,poria2017review}, 
and healthcare diagnosis~\cite{polsterl2021combining,wolf2022daft,narazani2022pet,hager2023best,wolf2023don}. 
Multimodal machine learning presents distinctive computational and theoretical research challenges due to the diversity of data sources involved~\cite{liang2021multibench,liang2023multiviz}. 
The impressive versatility and efficacy of multimodal neural network models can be attributed to their ability to effectively integrate and leverage heterogeneous data. 

Multimodal models process heterogeneous information obtained from multisource sensors by extracting their features. Subsequently, the extracted features are fused at different levels (including early, middle, or late fusion~\cite{wang2020deep,nagrani2021attention,narazani2022pet}) to address specific tasks, such as classification, recognition, description, and segmentation~\cite{liang2021multibench,liang2023multiviz,duquenne2021multimodal,akbari2021vatt,delpreto2022actionsense}. 
Heterogeneous information from multisource sensors, however, is susceptible to confounding effects caused by extraneous variables or multiple distributions~\cite{lu2021metadata,booth2021bias,belouali2021acoustic,NEURIPS2021_f740c8d9,wachinger2021detect,martin2022multimodality,bica2022transfer}. 
Confounding variables pertain to external factors that introduce bias (either positive or negative)  in the relationship between the variables being studied~\cite{pourhoseingholi2012control,soleymani2017survey}. The complexity of confounders emerges from their potential pervasiveness across diverse data modalities. For instance, in image analysis, confounders might encompass lighting variations, while in audio classification, speaker attributes like race or gender can be confounding factors. In video parsing, backgrounds play a role, and in dementia diagnosis, the education level of patients can be a confounder. Furthermore, positive or negative correlations can exist among heterogeneous data that impact the distributions of the learned features~\cite{soleymani2017survey,lu2021metadata}. 
These factors pose challenges for a multimodal network to accurately uncover the actual relationship between variables for a given task. 
Ignoring the confounding effects and partial dependencies in features or data can result in substantial drawbacks, including the deviation of a multimedia model from its global minimum, reduction in the speed and stability of neural network training, and the generation of unreliable or misleading outcomes. 
Consequently, it is recommended to normalize features from different modalities before their fusion~\cite{lu2021metadata,vento2022penalty}. In this study, we show that the normalization step can help mitigate the confounding effects and enhance the overall performance and reliability of the multimodal networks.

\begin{figure}
  \centering
        \includegraphics[width=1.\textwidth]{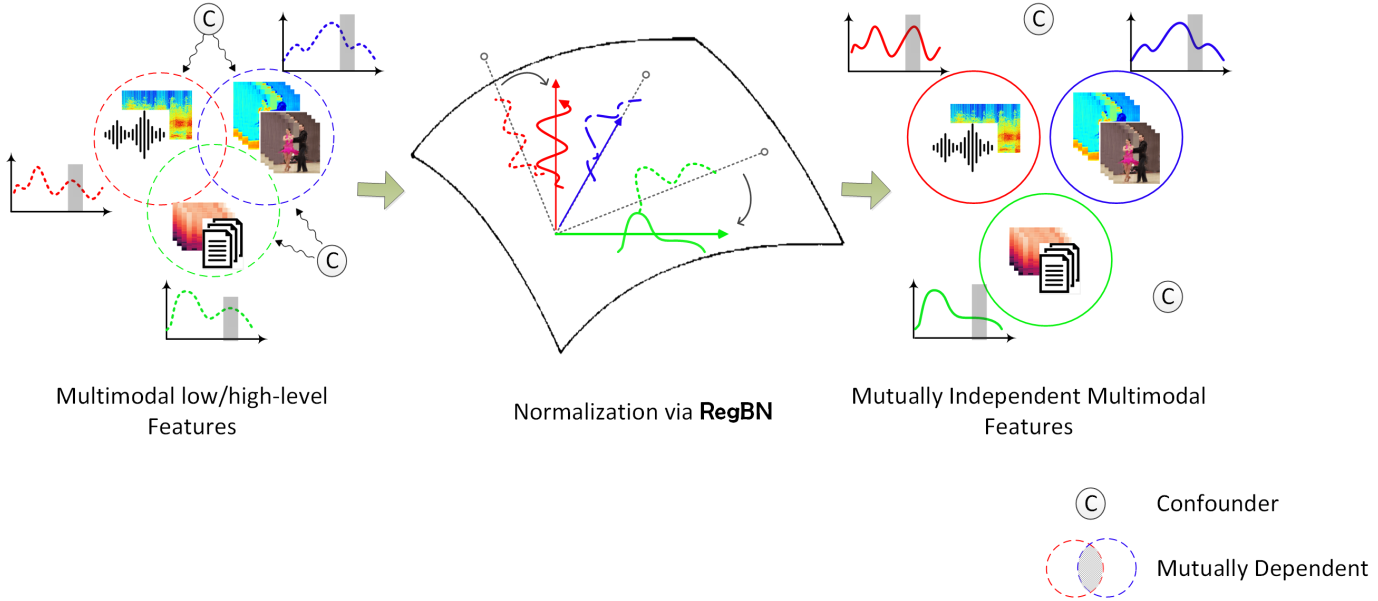}
\caption{The presence of heterogeneous data often entails external confounding effects (denoted by `C' in the figure) and partial dependencies, which impede the efficient training of a multimodal network. 
This study recommends the normalization of low- and high-level features extracted by foundation models through the use of RegBN.
This allows the normalised features to be rendered independent, enabling a multimodal network to discern underlying patterns and optimize its performance. 
}
\label{fig:framework}
\end{figure}


Normalization techniques have proven to be highly effective in enhancing the speed and stability of neural network training. By re-centering and re-scaling the input layers, the normalization methods provide a promising solution to overcome the challenges of training deep neural networks. 
Batch normalization (BN)~\cite{ioffe2015batch}, layer normalization (LN)~\cite{ba2016layer}, group normalization (GN)~\cite{wu2018group}, and instance normalization (IN)~\cite{wu2018group} are popular normalization techniques that have been used in the foundation of many state-of-the-art neural networks such as ResNet~\cite{he2016deep}, DenseNet~\cite{huang2017densely}, Inception~\cite{szegedy2016rethinking}, and others.  
The aforementioned normalization methods are designed to standardize feature distributions and do not take into account confounding effects and dependencies among features. 
Recent progress in multimodal learning leverages the potential of extensive multimodal data representations. CLIP~\cite{radford2021learning}, ALIGN~\cite{jia2021scaling}, and MaMMUT~\cite{kuo2023mammut} are multimodal foundations devised for images, text, audio, and video. Expanding this scope, ImageBind~\cite{girdhar2023imagebind} and Gato~\cite{reed2022generalist} tackle diverse tasks across multiple modalities. Dealing with large-scale multimodality data, however, may introduce new obstacles such as modality heterogeneity, modality imbalance, confounding, and intermodal variabilities. These emphasize the necessity of developing a normalization technique that is dedicated to multimodal data.

To tackle these challenges, several methods~\cite{lu2021metadata,vento2022penalty,zhao2020training} have recently been proposed that use metadata—the data that provides information about given data—for the normalization of the features in a neural network.
However, these studies predominantly focus on metadata and still encounter issues related to confounding effects. 
Motivated by these findings, we introduce a novel normalization method for multimodal heterogeneous data, referred to as RegBN, aimed at removing confounding effects and dependencies from low- and high-level features before fusing those. 
Our approach entails leveraging regularization to promote independence among heterogeneous data from multisource sensors (Figure~\ref{fig:framework}). RegBN facilitates the training of deep multimodal models while ensuring the prediction of reliable results. 
Our key contributions are as follows:
\begin{itemize}
\item As a normalization module, RegBN can be integrated into multimodal models of any architecture such as multilayer perceptrons (MLPs), convolutional neural networks (CNNs),  vision transformers (ViTs), and other architectures. 
\item RegBN possesses the capability to be applied to a vast array of heterogeneous data types, encompassing text, audio, image, video, depth, tabular, and 3D MRI. 
\item RegBN undergoes comprehensive evaluation on a collection of eight datasets, including multimedia, affective computing, healthcare diagnosis, and robotics.
\end{itemize}
The findings indicate that RegBN consistently leads to substantial improvements in inference accuracy and training convergence regardless of the data type or method used.

\section{Related work}
\textbf{Normalization methods in deep learning}: BN~\cite{ioffe2015batch}, LN~\cite{ba2016layer}, GN~\cite{wu2018group}, and IN~\cite{wu2018group} are conventional normalization methods that normalize the input features by re-centering and re-scaling. 
Unlike such methods that process input layers separately, RegBN takes two input layers and produces corresponding mutually independent output layers. The input layers could be any $n$-dimensional low-/high-level features, metadata, or raw data.

\textbf{Confounding effect removal}: Several statistical techniques have been developed for regressing out confounders~\cite{pourhoseingholi2012control,more2021confound,polsterl2021estimation}. Such methods are basically developed for confounding effect removal in healthcare. In recent years, several studies used deep neural networks for coping with confounders~\cite{zhao2020training,lu2021metadata,vento2022penalty}. 
\citet{lu2021metadata} introduced a novel layer normalization module called `metadata normalization' (MDN), which is designed specifically for neural networks that incorporate metadata. 
A closed-form solution to linear regression was developed by the authors to capture the relationship between feature layers and metadata. MDN, however, exhibits certain limitations such as its weak performance on small mini-batches and the requirement of the entire metadata for computing its matrix inverse. 
A penalty-based approach called PMDN \cite{vento2022penalty} was proposed as an improvement over MDN. PMDN addresses the shortcomings of MDN by using learnable parameters instead of estimating MDN's matrix inverse. PMDN is designed for metadata with a small number of features due to the huge number of required learning parameters. PMDN also requires a time-consuming two-step training procedure to optimize its learnable parameters. 
In addition, a major limitation of PMDN is how to effectively train the learnable parameters, as these parameters are trained solely based on the model's loss function, and models are typically unaware of the presence of confounding effects. 

We leverage the great potential of regularization as a solution to address the aforementioned issues. Our proposed normalization technique does not rely on learnable parameters and is capable of operating efficiently on small mini-batches. 
Most importantly, our approach produces more accurate results and is applicable to various types of multimodal neural networks, not limited to those incorporating metadata, and its ability to operate at both low- and high-feature levels. 
RegBN is the first normalization method to address the dependency and confounding issues in multimodal data, making it a unique and promising contribution to the field.

\section{RegBN: A regularization method for normalization of multimodal data}\label{sec:proposed}
Given a trainable multimodal neural network (e.g., MLPs, CNNs, ViTs) with multimodality backbones $\mathcal{A}$ and $\mathcal{B}$. 
Let $f^{(l)}\in{\mathcal{R}^{b\times n}}$ represent the $l$-th layer of the multimodal network for modality $\mathcal{A}$ with batch size $b$ and $n_1\times\ldots\times n_N$ features that are flattened into a vector of size $n$. 
In a similar vein, we define $g^{(k)} \in{\mathcal{R}^{b\times m}}$ as the $k$-th layer of the multimodal network for modality $\mathcal{B}$ with $m_1\times\ldots\times m_M$ features that are flattened into a vector of size $m$. 
Layers $l$ and $k$ can be positioned directly prior to the fusion step. 
Depending on the fusion approach employed, $f^{(l)}$ and $g^{(k)}$ can contain low-level features, high-level features, or latent representations of the multimodal model for early, middle, or late fusion, respectively (see Appendix~\ref{appendix:fusion}). 
Features $f^{(l)}$ and $g^{(k)}$ pertain to two distinct modalities and are subject to immeasurable unknown confounders. Moreover, they also share partial common information. 
The goal is to find potential similarities (mainly caused by confounding factors and dependencies during data collection) between these layers and then remove those (Figure~\ref{fig:framework}). Mutually-independent layers can be then fused in the multimodal network for a given task.  
To this end, one can represent  $f^{(l)}$ based on  $g^{(k)}$ as a linear regression model
\begin{equation}\label{eq:basic}
    f^{(l)} = W^{(l,k)} g^{(k)} + f^{(l)}_{r}.
\end{equation}
Here, $W^{(l,k)}$ is a projection matrix of size $n\times m$ and $f^{(l)}_{r}$ represents the difference between  $f^{(l)}$ and its corresponding map in the domain of $g^{(k)}$, also known as the residual. 
Ideally, the residual map does not contain any features of $g^{(k)}$, so $f^{(l)}_{r}$ and $g^{(k)}$ are mutually independent. 
RegBN minimizes the linear relationship between these layers via
\begin{equation}\label{eq:opt1}
    \hat{W}^{(l,k)}=\underset{W^{(l,k)}}{\mathrm{arg\,min}}\norm{f^{(l)}-W^{(l,k)} g^{(k)}}^2_2\;\;\; \textit{w.r.t.}\;\;\; \norm{W^{(l,k)}}_F=1.
\end{equation}
The constraint in terms of the Frobenius norm, $\norm{W^{(l,k)}}_F=\sum_{i=1}^{n}{\sum_{j=1}^m \omega^{(l,k)\:2}_{i,j}=1}$, guarantees that the network does not experience vanishing and exploding gradients during training and inference. 
Eq.~\ref{eq:opt1} is a regularization method for ill-conditioned problems~\cite{neumaier1998solving,calvetti2000tikhonov,moura2019stochastic,richards2021distributed}. 
A Lagrangian multiplier offers an equivalent formulation $\mathcal{F}$ to the optimization defined at Eq. \ref{eq:opt1}:
\begin{equation}
    \mathcal{F}(W^{(l,k)},\lambda_{+}) = \norm{f^{(l)}-W^{(l,k)} g^{(k)}}^2_2+\lambda_{+}
    \left(\norm{W^{(l,k)}}_F-1\right).
    \label{eq:lambda}
\end{equation}
In this equation, $\lambda_{+}$ is a positive Lagrangian multiplier, playing the role of a dual variable. Minimizing $\mathcal{F}(W^{(l,k)},\lambda_{+})$ over $W^{(l,k)}$ yields the projection matrix
\begin{equation}\label{eq:omega}
    \hat{W}^{(l,k)} = \left(g^{(k)\top} g^{(k)}+\hat{\lambda}_{+}\mathbf{I}\right)^{-1}g^{(k)\top} f^{(l)},
\end{equation} 
where, $\hat{\lambda}_{+}$ is the estimated Lagrangian multiplier obtained through minimization of the term $\norm{W^{(l,k)}}_F-1$; $\mathbf{I}$ is an identity matrix and superscript $\top$ represents the transpose operator. We employ singular value decomposition (SVD) and limited-memory Broyden-Fletcher-Goldfarb-Shanno (L-BFGS)~\cite{liu1989limited} for solving Eq.~\ref{eq:opt1}. The solution is detailed in Appendix~\ref{appendix:sol}, and the proposed method is summarized in Algorithm~\ref{alg:1}. 
It is worth noting that the projection weights are not directly trained by the multimodal criterion since there is a danger of being fooled by confounders. Instead, the projection weights are learned through Eqs.~\ref{eq:opt1}-\ref{eq:omega} and updated recursively over chunks of training data, detailed below. 

\begin{figure}
  \centering
    \begin{subfigure}[t]{0.45\textwidth}
        \centering
        \includegraphics[height=0.75\textwidth]{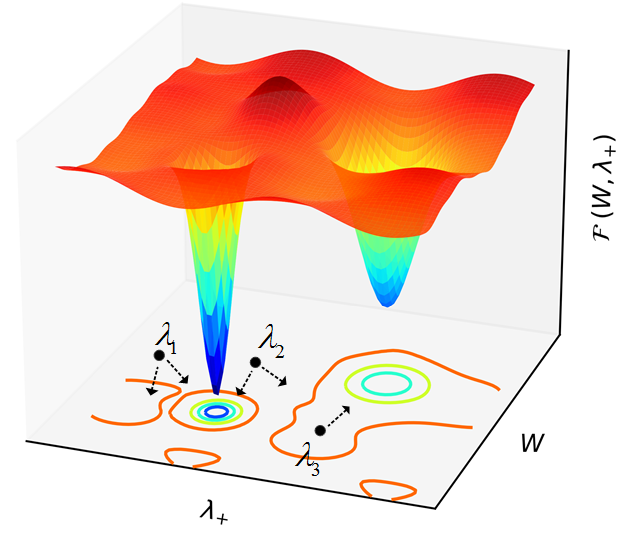}
        \caption{Extrema in a mini-batch and seeded values $\{\lambda_1,\lambda_2,\lambda_3\}$ for finding sub-optimal $\lambda_+$}
        \label{fig:local}
    \end{subfigure}%
    \hspace{1em}
    \centering
    \begin{subfigure}[t]{0.45\textwidth}
        \centering
        \includegraphics[height=0.7\textwidth]{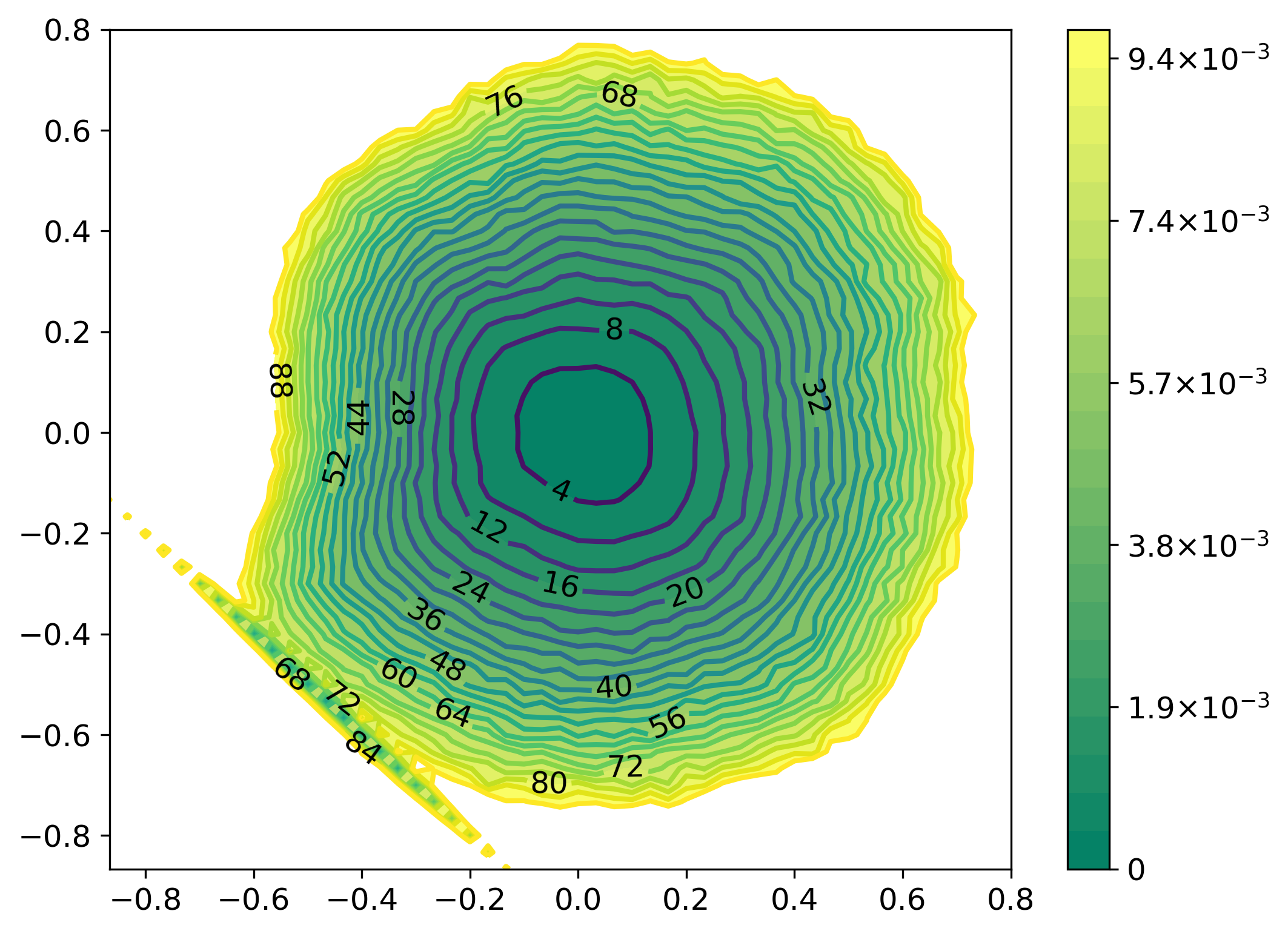}
        \caption{2D visualization of a multimodal LeNet model with RegBN on AV-MNIST~\cite{ma2021smil}}
        \label{fig:landscape}
    \end{subfigure}
\caption{(a) In a mini-batch, there is a risk of falling into local minima; we fix this problem by seeding $\lambda_{+}$ at different locations, detailed in Section~\ref{sec:avoidextrema}.
(b) 2D visualization of the loss surface of the multimodal SMIL model~\cite{ma2021smil} on AV-MNIST (see Section~\ref{sec:multimedia}). The numbers inside the plot indicate the classification loss, while the color bar reports the L1 loss of the projection weights in RegBN, i.e. $\Delta W_t$.  RegBN's weights converge into their global minimum as long as the multimodal model reaches its global minimum.
}

\end{figure}

\begin{algorithm}
	\caption{Pseudocode of RegBN} 
	\begin{algorithmic}[1]
     \State \textbf{Inputs}: $f^{(l)}\in{\mathcal{R}^{b\times n}}$ (the flattened $l$-th layer with $n$ learnable/non-learnable features and batch size $b$), $g^{(k)}\in{\mathcal{R}^{b\times m}}$ (the flattened $k$-th layer with $m$ learnable/non-learnable features), 
     $t$ (timestep/mini-batch), 
     $W^{(l,k)}_{t-1}$ (the projection weights of the previous timestep), 
     $\Lambda_{t-1}$ (a collection of estimated $\lambda_+$ values until timestep `$t-1$'), 
     $\gamma_t$ (the learning rate  of the multimodal model at timestep $t$), \emph{training};
    \State Compute SVD of $g^{(k)}$: SVD$\bigl(g^{(k)} \bigr) = U \Sigma V^* = \sum_{i=1}^m \sigma_{i} {u_i} {v^*_i}$
    \State Set $\mathbf{\Lambda}_{t}$ through Eq.~\ref{eq:lambdai} and given $\Lambda_{t-1}$;
 \For {$\lambda$ in $\mathbf{\Lambda}_{t}$}
    \State Initialize ${\lambda}_+$ with $\lambda$ and then apply the L-BFGS algorithm to approximate Eq.~\ref{eq:lambdahat};
    \State Store the estimated ${\lambda}_+$ and the approximation error;
	\EndFor
    \State Choose ${\lambda}_+$ with the lowest approximation error as the sub-optimal $\hat{\lambda}_+$;
    \State Update $\Lambda_t$: $\Lambda_t\leftarrow \Lambda_{t-1} \cup \hat{\lambda}_t$;
    \State Insert $\hat{\lambda}_+$ to Eq.~\ref{eq:omega} for computation of the projection matrix: $\hat{W}^{(l,k)}=\sum_{i=1}^{m}{\bigl(\frac{\sigma_{i} {u_i} f^{(l)}}{\sigma_{i}^2+\hat{\lambda}_+}\bigr)v_i}$;
	\If {\emph{training}}
    \State Normalize the input feature layer using $\hat{W}^{(l,k)}$: $f^{(l)}_{r} \leftarrow f^{(l)} - \hat{W}^{(l,k)} g^{(k)}$;
	\EndIf
    \State Update the projection weights $W^{(l,k)}_{t}$ via Eqs.~\ref{eq:w1} and \ref{eq:w2};
	\If {\emph{not training}}
    \State Normalize the input feature layer using $W^{(l,k)}_{t}$: $f^{(l)}_{r} \leftarrow f^{(l)} - W^{(l,k)}_{t} g^{(k)}$;
	\EndIf\\
    \Return $f^{(l)}_{r}$, $W^{(l,k)}_{t}$, and $\Lambda_t$
	\end{algorithmic} 
\label{alg:1}
\end{algorithm}
\subsection{Update of the projection matrix}\label{sec:updateW}

The projection matrix, ${W}^{(l,k)}$, is calculated for every mini-batch so we recursively update that via the exponential moving average's approach~\cite{kingma2014adam}, which decays the mean gradient and variance for each variable exponentially. Let $W^{(l,k)}_{t-1}$ denote the updated projection matrix until the previous timestep.  
We define $\Delta W_t$ as the mean absolute error between the currently predicted projection matrix and the previously updated projection matrix: $\Delta W_t=\norm{\hat{W}^{(l,k)}- W^{(l,k)}_{t-1}}_1$. The projection matrix is then updated at timestep $t$ via
\begin{equation}\label{eq:w1}
    W^{(l,k)}_{t} = \left(1-\gamma_t\frac{m_t}{\sqrt{\nu_t+\epsilon}}\right) \hat{W}^{(l,k)}+ \gamma_t\frac{ m_t}{\sqrt{\nu_t+\epsilon}} W^{(l,k)}_{t-1}.
\end{equation} 
Here, $\gamma_t$ is the learning rate of the multimodal model at timestep $t$; $m_t$ and $\nu_t$ are the first and second moments, respectively, derived from 
\begin{equation}\label{eq:w2}
    m_t = \frac{\beta_1 m_{t-1}+(1-\beta_1) \Delta W_t}{1-\beta_1^t}, \:
    \nu_t = \frac{\beta_2 \nu_{t-1}+(1-\beta_2) \Delta W_t^2}{1-\beta_2^t},
\end{equation} 
where $\beta_1, \beta_2 \in (0,1)$ represent constant exponential decay rates. As mentioned in Algorithm~\ref{alg:1}, the projection matrix is updated during training only (Algorithm~\ref{alg:1}: Steps 11-13), and we use the latest updated projection weights for validation or inference (Algorithm~\ref{alg:1}: Steps 15-17).

\subsection{Avoiding falling into local minima}\label{sec:avoidextrema}

Multimodal networks often contain several backbones (e.g., MLPs, CNNs, ViTs), depending on the number of multiple heterogeneous data sources.
Identifying the global minima of such networks is a challenging problem, as the networks may fall into local minima. 
On the other hand, optimization of Eq.~\ref{eq:lambda} relies on ${\lambda}_{+}$, which may introduce several local minima. 
Conventional regression methods treat $\lambda_{+}$ as a constant/hyperparameter, while predetermined hyperparameters usually increase the risk of falling into local minima. 
To prevent this, we adopt a mini-batch-wise approach for the estimation of $\lambda_{+}$. 
The objective is to estimate suboptimal $\lambda_{+}$ with the L-BFGS optimization algorithm in a way that meets
\begin{equation}\label{eq:lambdahat}
    \hat{\lambda}_{+} = \underset{{\lambda}_{+}}{\mathrm{arg\,min}}\frac{\partial \mathcal{F}(W^{(l,k)},\lambda_{+})}{\partial \lambda_{+}} = 
    \underset{{\lambda}_{+}}{\mathrm{arg\,min}} \left(\norm{\bigl(g^{(k)\top} g^{(k)}+{\lambda}_{+}\mathbf{I}\bigr)^{-1}g^{(k)\top} f^{(l)}}_F-1\right).
\end{equation} 
Quasi-Newton methods like L-BFGS rely on the initial value of ${\lambda}_{+}$ for minimization of the optimization problems like Eq.~\ref{eq:lambdahat} (see Figure~\ref{fig:local}). 
Due to the limited number of local minima that exist per mini-batch, we adopt a method whereby we initialize $\lambda_{+}$ with multiple random values within a predetermined range $\Lambda_p=[\lambda_{p_1},\ldots, \lambda_{p_C}]$ as well as with the median of the suboptimal $\lambda_{+}$ values predicted in the previous mini-batches, $\mu_{\frac{1}{2}}({\Lambda}_{t-1})$:
\begin{equation}\label{eq:lambdai}
    \mathbf{\Lambda}_{t} = \Lambda_p\cup \mu_{\frac{1}{2}}({\Lambda}_{t-1}) = [\lambda_{p_1},\ldots, \lambda_{p_C}] \cup \mu_{\frac{1}{2}}\{[\hat{\lambda}_{1},\ldots,\hat{\lambda}_{t-1}]\}.
\end{equation} 
By creating a loop over $\mathbf{\Lambda}_{t}$, we first initialize $\lambda_{+}$ and then estimate its true value using the L-BFGS method. We also store the approximation error and after the termination of the loop, the suboptimal value of $\lambda_{+}$ for a given mini-batch is determined by selecting the estimated value that yields the lowest approximation error.  
After calculating the suboptimal lagrangian multiplier, we update $\Lambda$, i.e., $\Lambda_t\leftarrow \Lambda_{t-1} \cup \hat{\lambda}_t$. 
As illustrated in Figure~\ref{fig:landscape}, when the multimodal model approaches its global minimum, the parameters of RegBN also converge.
\section{Experiments}

Heterogeneous data captured by diverse multisource sensors are utilized to verify the usefulness and effectiveness of RegBN in multiple data contexts such as language, audio, 2D image, video, depth, 3D MRI, and tabular data. RegBN is applied to eight datasets that are summarized in Table~\ref{table:datasets}. 
Details on the datasets and baseline methods are provided in Appendices~\ref{appendix:datasets} \& \ref{appendix:methods}. 
The default parameters and settings for RegBN are reported in Appendix~\ref{appendix:regbn}. Our code is openly available at \url{https://mogvision.github.io/RegBN}. Experimental details are provided in Appendix~\ref{appendix:results}. Here, we summarize the main results.
The findings of this investigation provide insights into the performance of RegBN across various data modalities and highlight its potential as a robust normalization technique. 
PMDN relies on a considerable number of learnable parameters specifically tailored for metadata with limited dimensions. MDN necessitates substantial RAM resources, particularly when estimating its inverse matrix on large-scale datasets.

\begin{table}
  \caption{The experimental section covers a diverse range of research areas, dataset
sizes, input modalities (in the form of $i$: image, $l$: language, $v$: video, $a$: audio, $s$: 3D snippet-level features, $m$: 3D MRI, $d$: depth, $t$: tabular, $f$: force sensor, $p$: proprioception  sensor), and prediction tasks.}
  \label{table:datasets}
  \centering
  {\setlength{\tabcolsep}{3.5pt}
  \begin{tabular}{@{\extracolsep{4pt}}lccrrc@{}}
    \toprule
     Area  & Dataset & Modalities & \multicolumn{2}{c}{Samples (\#)}  & Prediction task \\
    \cmidrule{4-5} 
    &&&Training& Test\\ 
    \midrule
    Affective computing & CMU-MOSEI& $\{v,a,l\}$ & 18,118 & 4,659 & sentiment\\
    Affective computing & CMU-MOSI & $\{v,a,l\}$ & 1,513  & 686   & sentiment\\
    Affective computing & IEMOCAP  & $\{v,a,l\}$ & 3,515 & 938   & emotion\\
    Multimedia & MM-IMDb  & $\{i,l\}$   & 18,160 & 7,799 & movie genre\\
    Multimedia & LLP      & $\{a,v,s\}$ & 10,649 & 1,200 & video parsing\\
    Multimedia & AV-MNIST (small) & $\{i,a\}$   & 1,045  & 450   & classification\\
    Multimedia & AV-MNIST & $\{i,a\}$   & 60,000  & 10,000   & classification\\
    Healthcare diagnosis  & ADNI & $\{m,t\}$ & 1,073 & 268 & dementia diagnosis\\
    Robotics & Vision\&Touch & $\{i, d, f, p\}$ & 40,546 & 22,800 & contact\\
    - &  Synthetic dataset & $\{i,t\}$ & 10,000 & 1,000 & classification\\
    \bottomrule
  \end{tabular}
  }

\end{table}
\subsection{Multimedia}\label{sec:multimedia}

\textbf{Experiments with the LLP dataset~\cite{tian2020unified}}:
Audio-Visual Video Parsing (AVVP)~\cite{tian2020unified} (see Appendix~\ref{appendix:AVVP}) is employed as a baseline for parsing individual audio, visual, and audio-visual events under both segment-level and event-level metrics over the LLP dataset. 
The normalization module is employed to decouple the audio features from the video features, ensuring their independence. 
Table~\ref{table:llp} reports that RegBN improves the baseline performance in nine out of ten metrics.
RegBN brings about improvements in all audio-visual video parsing subtasks, measured both at the segment-level and event-level metrics. These findings suggest that decoupling audio and video allows AVVP to produce more accurate predictions of event categories on a per-snippet basis.\footnote{Appendix~\ref{appendix:multimedia} reports the validation results of AVVP with and without RegBN on the LLP dataset.}

\textbf{Experiments with the MM-IMDb dataset~\cite{arevalo2017gated}}:
The MM-IMDb dataset was curated for the purpose of predicting movie genres through the use of either image or text modality. This task involves multi-label classification since a single movie may be associated with multiple genres (see Appendix~\ref{appendix:MM-IMDb}). 
SMIL~\cite{ma2021smil}, as a baseline approach, employs the pre-trained BERT to extract the textual features, while image features are extracted using the pre-trained VGG-19. The text features are subsequently normalized with respect to the visual features using the proposed RegBN before the fusion process occurs. The fusion operation involves a concatenation layer. The fused features are passed through two fully-connected layers to obtain the classification labels. 
The details are provided in Appendix~\ref{appendix:SMIL}. 
Table~\ref{table:SMIL} presents the quantitative results obtained from the experiments. The table indicates that applying a normalization layer such as PMDN or RegBN can lead to an improvement in the performance over the baseline model. This highlights the importance of the normalization step in multimedia data. In particular, RegBN was found to enhance the F1 score of the baseline by an average of 8.86\% across all three metrics.

\begin{figure*}
    \centering
    \begin{subfigure}[t]{0.19\textwidth}
        \centering
        \includegraphics[height=0.8\textwidth]{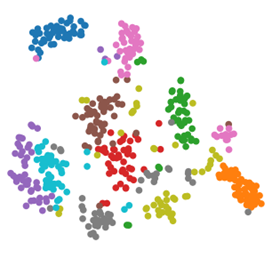}
        \caption{Only image}
    \end{subfigure}%
    \begin{subfigure}[t]{0.19\textwidth}
        \centering
    \includegraphics[height=0.8\textwidth]{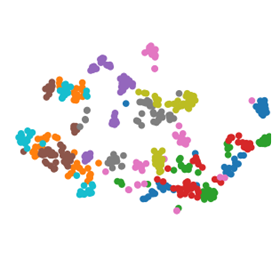}
    \caption{Only spectrogram}
    \end{subfigure}
    \begin{subfigure}[t]{0.19\textwidth}
        \centering
        \includegraphics[height=0.8\textwidth]{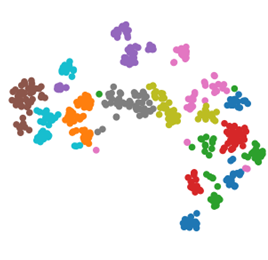}
        \caption{MDN}
    \end{subfigure}
    \begin{subfigure}[t]{0.19\textwidth}
        \centering
        \includegraphics[height=0.8\textwidth]{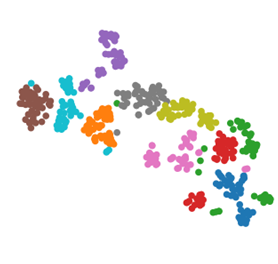}
        \caption{PMDN}
    \end{subfigure}
    \begin{subfigure}[t]{0.19\textwidth}
        \centering
        \includegraphics[height=0.8\textwidth]{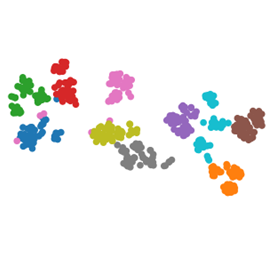}
        \caption{RegBN}
    \end{subfigure}\\
    \includegraphics[height=0.06\textwidth]{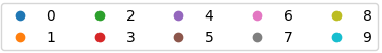}

    \caption{tSNE visualization of the features extracted from a-b) an unimodal image and an unimodal audio, and c-e) the multimodal model with different normalization methods. Each data point represents a sample.
}    
\label{fig:smallAVmnist}
\end{figure*}

\textbf{Experiments with the AV-MNIST dataset~\cite{ma2021smil}}:
The small AV-MNIST dataset comprises two modalities, namely, audio and image. The audio modality is obtained from Free Spoken DigitsDataset\footnote{Free Spoken Digit Dataset is available at \url{https://github.com/Jakobovski/free-spoken-digit-dataset}}, which includes 1,500 raw audio recordings from three different speakers. Despite the independence of the image and audio modalities, the primary objective of this experiment is to investigate whether the confounding effects induced by the speakers have an impact on the classification results. 
In line with the previous experiment, SMIL~\cite{ma2021smil} serves as the baseline method. The obtained classification results for the combinations of baseline+MDN, baseline+PMDN, and baseline+RegBN are 98.43\%, 98.68\%, and 99.11\%, respectively. 
Figure~\ref{fig:smallAVmnist} illustrates t-SNE plots for the different approaches. 
In this experiment, tSNE is applied to the features extracted from a fully-connected layer after the concatenation of normalized image features with audio features. The figure demonstrates that the classes are more separable for RegBN as compared to MDN or PMDN, in particular, class `2'. This result implies that our normalization method effectively removes the confounding effects caused by different speakers, thereby providing more flexibility to the baseline method for accurate classification. The results on AV-MNIST are reported in Appendix \ref{appendix:multimedia}.

\begin{table}
  \caption{Audio-visual video parsing accuracy (\%) of AVVP~\cite{tian2020unified}, as baseline (BL), on the LLP dataset~\cite{tian2020unified} for different normalization techniques. A and V stand for audio and visual, respectively.}
  \label{table:llp}
  \centering
  {\setlength{\tabcolsep}{4.2pt}
  \begin{tabular}{@{\extracolsep{4pt}}l ccccc ccccc@{}}
    \toprule
    Method  & \multicolumn{5}{c}{Segment-Level} &  \multicolumn{5}{c}{Event-Level} \\
    \cline{2-6} \cline{7-11} \addlinespace
    & A$\uparrow$ & V$\uparrow$  & A-V$\uparrow$  & Type$\uparrow$ & Event$\uparrow$ & A$\uparrow$ & V$\uparrow$  & A-V$\uparrow$  & Type$\uparrow$ & Event$\uparrow$\\
    \midrule
    BL       & 60.1 & 52.9 & {48.9} & {54.0} & {55.4} & 51.3 & 48.9 & 43.0 & 47.7 & 48.0\\
    BL+PMDN  & 59.7 & 53.0 & 48.8 & 53.6 & 55.3 & 51.4 & 49.1 & 42.9 & 47.7 & 48.2\\
    BL+RegBN  & {60.2} & {53.3} & {48.9} & {54.0} & 55.2 & {52.0} & {49.5} & {43.1} & {47.9} & {49.3}\\
    \bottomrule
  \end{tabular}
  }
\end{table}

\begin{table}
  \caption{ Multi-label classification scores (F1 score) of baseline SMIL~\cite{ma2021smil} (denoted by BL) with/without normalization on the MM-IMDb dataset~\cite{arevalo2017gated}}
  \label{table:SMIL}
  \centering
  \begin{tabular}{@{\extracolsep{4pt}}l c ccc@{}}
    \toprule
    Method  &  Norm. Params.& \multicolumn{3}{c}{F1 Score (\%)} \\ 
    \cline{3-5} \addlinespace & ($\#$) & Samples$\uparrow$ & Micro$\uparrow$  & Weighted$\uparrow$ \\
    \midrule
    BL        & -- & 49.62 & 51.18  & 48.95\\
    BL+PMDN   & 65,536 &  49.85 & 51.44 & 49.36\\
    BL+RegBN  & 0 & {54.82} & {55.37} & {52.83}\\
    \bottomrule
  \end{tabular}
\end{table}

\subsection{Affective computing}\label{sec:affective}

In this section, we explore the potential application of RegBN in multimodal emotion-sentiment analysis using the IEMOCAP~\cite{busso2008iemocap}, CMU-MOSI, and CMU-MOSEI~\cite{zadeh2016mosi} datasets, which are divided into aligned and non-aligned multimodal time-series (see Appendices~\ref{appendix:IEMOCAP}, \ref{appendix:MOSI}, \& \ref{appendix:MOSEI}). The baseline method (BL) for this task is the Multimodal Transformer (MulT)~\cite{tsai2019multimodal}, which is a ViT developed for analyzing multimodal language sequences. MulT fuses video ($v$), audio ($a$), and language ($l$) at three levels ({see Appendix~\ref{appendix:MulT}}): 1) language fusion ($v\rightarrow l$ \& $a\rightarrow l$), 2) audio fusion ($v\rightarrow a$ \& $l\rightarrow a$), and 3) video fusion ($a\rightarrow v$ \& $l\rightarrow v$). The outputs of one or more levels are then aggregated for classification. Here, we report the results of all three fusion levels combined, while the detailed results for each level are provided in Appendix~\ref{appendix:affective}. 
The quantitative results of the emotion analysis experiment are reported in Table~\ref{table:Affective-IEMOCAP}. The table shows that RegBN improves the classification performance of MulT in most cases, particularly for the ``natural'' class of the word-aligned experiment. 
Table~\ref{table:Affective-CMU} presents the results of the multimodal sentiment analysis on the CMU-MOSI and CMU-MOSEI datasets. In both the word-aligned and unaligned experiments, decoupling the multimodal features with RegBN improves training and inference performance. 

\subsection{Healthcare diagnosis}\label{sec:diagnosis}

The ADNI dataset~\cite{jack2008alzheimer} includes 3D MRI scans of patients along with rich clinical information organized in a low-dimensional tabular format (see Appendix~\ref{appendix:ADNI}). The dataset is subject to confounding effects such as age, sex, and level of education, which must be considered to prevent biased evaluation results, as noted in studies~\cite{wen2020convolutional,polsterl2022identification}. Similar to \cite{wolf2022daft}, we partitioned the data into five separate and non-overlapping folds, ensuring that each fold had a balanced distribution of diagnosis, age, and sex.  
In this experiment, we utilized a 3D ResNet~\cite{wolf2022daft} for extracting MRI features, while tabular features were extracted using an MLP network proposed in~\cite{hager2023best} (see Appendix~\ref{appendix:MMCL-TI}). These features were then concatenated and fed into another MLP network for diagnosis classification. We applied different normalization techniques to the unimodal MRI and tabular features prior to concatenation. 
Table~\ref{table:dem} reports the mean and standard deviation obtained by averaging the performance of the multimodal over these five folds. The complete results can be found in Appendix~\ref{appendix:dem}. 
RegBN emerged as the most effective technique, outperforming other normalization methods in terms of both ACC and BA metrics. Moreover, RegBN's ability to relax the interdependencies between the different modalities facilitated lower training loss in the multimodal network.

\begin{table}
  \caption{Results of multimodal emotion analysis on IEMOCAP~\cite{busso2008iemocap} with word aligned multimodal sequences. The baseline (BL) is Multimodal Transformer (MulT)~\cite{tsai2019multimodal}.}
  \label{table:Affective-IEMOCAP}
  \centering
  {\setlength{\tabcolsep}{4.5pt}
  \begin{tabular}{@{\extracolsep{4pt}}l cc cc cc cc cc@{}}
    \toprule
    Method  & \multicolumn{2}{c}{Loss} & \multicolumn{2}{c}{Happy (\%)} & \multicolumn{2}{c}{Sad (\%)} & \multicolumn{2}{c}{Angry (\%)} & \multicolumn{2}{c}{Natural (\%)} \\
    \cmidrule{2-3}  \cmidrule{4-5}  \cmidrule{6-7}  \cmidrule{8-9}  \cmidrule{10-11}  
    & Train$\downarrow$ & Test$\downarrow$ & Acc$\uparrow$ & F1$\uparrow$ & Acc$\uparrow$ & F1$\uparrow$ & Acc$\uparrow$ & F1$\uparrow$ & Acc$\uparrow$ & F1$\uparrow$\\
    \midrule
    BL  & 0.106 & 0.536  & 86.3 & {84.0}  & 81.5  & 80.6 & 86.5 & 86.4 & 69.5 & 69.1\\
    BL+RegBN  & {0.009} & {0.452}  & {87.4} & 83.0& {84.3}  & {84.1} & {88.2} & {88.1} & {73.4} & {73.2}\\
    \bottomrule
  \end{tabular}
  }
\end{table}
\begin{table}
  \caption{Multimodal sentiment analysis results multimodal on  CMU-MOSEI \& MOSI~\cite{zadeh2016mosi} with word aligned multimodal sequences. The baseline (BL) method is MulT~\cite{tsai2019multimodal}.}
  \label{table:Affective-CMU}
  \centering
  {\setlength{\tabcolsep}{4.5pt}
  \begin{tabular}{@{\extracolsep{4pt}}l c cc ccccc@{}}
    \toprule
    Method   & Dataset & \multicolumn{2}{c}{Loss} & \multicolumn{5}{c}{Sentiment (\%)} \\
    \cmidrule{3-4}  \cmidrule{5-9} 
    & &Training$\downarrow$ & Test$\downarrow$ & Acc$_2$$\uparrow$ & Acc$_5$$\uparrow$  & Acc$_7$$\uparrow$ & F1$\uparrow$ & Corr$\uparrow$\\
    \midrule
    BL &  CMU-MOSEI & 0.452 & 0.636 & 80.3 & 51.9 & 50.3  & 80.1  & {67.1}\\
    BL+RegBN  & CMU-MOSEI & {0.438} & {0.611} & {81.1} & {52.2} & {50.5}  & {81.2}  & 66.6\\
    \hdashline \addlinespace
        BL   &  CMU-MOSI &0.403 & 0.632 & 81.4 & {42.5}  &  37.5 & 82.0  & {69.7}\\
    BL+RegBN &CMU-MOSI  & {0.267} & {0.546} & {81.8} & 42.3  & { 38.6} & {82.3}  & 69.1\\
    \bottomrule
  \end{tabular}
  }
\end{table}

\begin{table}
  \caption{Training's cross-entropy (CE) loss along test's accuracy (ACC) and balanced accuracy (BA) results on the ADNI dataset~\cite{jack2008alzheimer}. Baseline is a combination of techniques developed in~\cite{hager2023best,wolf2022daft}. 
  }
  \label{table:dem}
  \centering
  \begin{tabular}{@{\extracolsep{4pt}}lcccc@{}}
    \toprule
     Method  &Norm. Params. & Training CE  & \multicolumn{2}{c}{Test}  \\
    \cmidrule{4-5} 
    & ($\#$) &loss$\downarrow$ & ACC (\%) $\uparrow$ & BA (\%)$\uparrow$ \\ 
    \midrule
    BL+BN    & 512 &  0.641±0.01  &  48.8±4.4 & 48.3±4.5\\
    BL+MDN   & 0 & 0.619±0.03 & 50.4±4.8 & 50.1±5.6 \\
    BL+PMDN  & 4,096 & 0.632±0.03 & 49.7±6.2 & 49.6±7.5 \\
    BL+RegBN & 0 & {0.596±0.01} & {53.0±3.1} & {52.3±3.7} \\
    \bottomrule
  \end{tabular}
\end{table}

\subsection{Robotics}\label{sec:robotics}
The Vision\&Touch dataset~\cite{lee2019making} encompasses recordings of simulated and real robotic arms that are equipped with visual (RGB and depth), force, and proprioception sensors (see Appendix~\ref{appendix:Vision&Touch}). We employed `Making Sense of Vision and Touch' (MSVT) developed by \citet{lee2019making} as a baseline (refer to Appendix~\ref{appendix:MSVT}).
For this experiment, we apply normalization techniques to decouple the visual RGB and force from depth information. Table \ref{table:TouchVision} summarizes the results, while detailed results can be found in Appendix \ref{appendix:robotics}. 
The flow loss values in these tables indicate that MSVT converges more effectively when used in conjunction with RegBN. This suggests that RegBN is able to effectively remove dependencies between the RGB and depth pair, as well as the force and depth pair. 

\begin{table}
  \caption{Results of MSVT~\cite{lee2019making} incorporating various normalization methods on Vision\&Touch~\cite{lee2019making}}
  \label{table:TouchVision}
  \centering
  \begin{tabular}{@{\extracolsep{4pt}}l cccc@{}}
    \toprule
     Method  & Norm. Params.   & \multicolumn{2}{c}{Training} & Test Accuracy \\
    \cmidrule{3-4} 
    &  ($\#$) & Flow loss$\downarrow$ & Total loss$\downarrow$ & (\%)$\uparrow$  \\
    \midrule
    BL     & -- &  0.212 & 0.563 &  86.2\\
    BL+PMDN  & 131,072 & 0.194 &  0.454 & 87.9 \\
    BL+RegBN  & 0 & {0.052} & {0.231} & {91.5}\\
    \bottomrule
  \end{tabular}
\end{table}

\begin{table}
  \caption{Classification accuracy results on the synthetic dataset in the presence of a confounder. Due to the randomness of the experiment, we reported the mean and std of results over 100 runs.}
  \label{table:synth}
  \centering
  \begin{tabular}{@{\extracolsep{4pt}}l cc cccc@{}}
    \toprule
    Normalization   &\multicolumn{2}{c}{Norm. params. ($\#$)}   & \multicolumn{2}{c}{Experiment I} & \multicolumn{2}{c}{Experiment II}  \\
    \cmidrule{2-3} \cmidrule{4-5}  \cmidrule{6-7} 
    Method &  MLP & CNN & MLP & CNN & MLP & CN \\
    \midrule
    \emph{reference} &  -- &   --    & $87.5$ & $87.5$  & $75.0$& $75.0$\\
    BN   &   256   &   112         & 96.4±0.4 & 94.9±0.2 & 83.1±0.8  & 85.2±0.3\\
    MDN   &   0   &   0            & 91.9±0.7 & 89.7±0.7  & 79.1±0.6 & 82.2±0.7\\
    PMDN  &  2048 &    201,728   &   93.4±0.6 &   92.7±0.5 &   80.8±0.7  &   84.3±0.8\\
    RegBN &   0 &   0              & 87.3±0.9 & 88.2±0.8  & 76.2±0.8 & 76.8±0.9\\
    \bottomrule
  \end{tabular}
\end{table}

\subsection{Synthetic dataset}\label{sec:synth}
Drawing from the synthetic dataset in study~\cite{lu2021metadata}, we designed a synthetic experiment for a binary classification task. 
Images with a size of 64$\times$64 are split into four regions with identical Gaussians in their background. The regions along the main diagonal are linked to the class label, while the off-diagonal ones act as confounders that do not impact the true label. 
We created metadata of size 16 for this dataset that includes the true binary label, the confounder value, one fake label, 12 randomly generated fake features, and 1 column with a value of one. 
The binary classes have overlapping, so the accuracy drops based on the amount of overlapping (see Appendix~\ref{appendix:synth}).  
In Experiment I, the theoretical maximum accuracy of classification is 87.5\%, while for Experiment II, it is 75\%. The base networks employed here are a simple CNN and an MLP-based neural network. The dataset and base networks are detailed in Appendices~\ref{appendix:synth} and~\ref{appendix:CNNandMLP}, respectively. 
The classification results are presented in Table~\ref{table:synth}, with more detailed results available in Appendix~\ref{appendix:synthResults}. The results reveal that conventional normalization techniques, such as BN, GN, and IN, are unable to remove the confounding effects. 
The performance of the base networks that employed MDN was superior compared to those with conventional normalization methods. However, MDN still exhibits shortcomings in dealing with confounding effects. 
Despite the incorporation of a high number of learnable parameters, PMDN was unsuccessful in eliminating confounding effects, as the inclusion of fake values in the metadata led to the deception of the multimodal model. 
RegBN demonstrated its ability to remove confounder effects by yielding performance closest to the theoretical maximum in both experiments. 

\subsection{Computational cost and ablation study}\label{sec:abalation}
We employed an NVIDIA GTX 1080 Ti with 12GB VRAM for image experiments and an NVIDIA A100 with 80GB VRAM for video experiments. Since RegBN learns the projection matrix through Eqs.~\ref{eq:omega}-\ref{eq:lambdahat}, it eliminates the need for learnable parameters, resulting in enhanced computational efficiency for multimodal models. 
During the training process, RegBN achieves a frame rate of approximately 124 fps when normalizing a pair of layers with dimensions $256\times1024$. In comparison, conventional normalization techniques achieve a frame rate of around 13,000 fps for the normalization of one layer of dimensions $256\times1024$. 
Regarding inference time, RegBN exclusively relies on the updated projection matrix and therefore does not require additional computation. 
RegBN is frequently used before the fusion blocks in multimodal models, so the training time of a few RegbBN layers is negligible. 
Results for the ablation study are provided in Appendix~\ref{appendix:abalation}.


\section{Conclusion and future work}
This study presented a novel normalization method designed for dependency and confounding removal in multimodal data and models. 
The experimental results demonstrated the effectiveness of RegBN in significantly enhancing the accuracy and convergence of multimodal models, rendering it a promising normalization method for multimodal heterogeneous data. 
The current version of RegBN employs L-BFGS on a single GPU, which will be improved in the future. 
Our aspiration is that this work effectively leverages new opportunities in exploring and harnessing multi-modality data within multimodal analysis.

\section*{Acknowledgements}
This work was supported by the Munich Center for Machine Learning (MCML) and the German Research Foundation (DFG). 
The authors gratefully acknowledge the computational and data resources provided by the Leibniz Supercomputing Centre. 
The authors thank Emre Kavak for his valuable input on the synthetic dataset design.

\newpage
\bibliographystyle{plainnat}
\bibliography{bibliography}

\newpage
\addtocontents{toc}{\protect\setcounter{tocdepth}{2}}

\setcounter{section}{0}
\renewcommand{\thesection}{\Alph{section}}


\begin{center}
    \begin{Large} {  RegBN: Batch Normalization of Multimodal Data with Regularization \\ (Supplementary Material)}
    \end{Large}%
\end{center}
\tableofcontents
\section{Multimodal normalization and fusion}\label{appendix:fusion}
\begin{figure}
\begin{center}
\captionsetup{justification=centering}
     \begin{subfigure}{1\columnwidth}
         \centering
         \includegraphics[height=0.2\textheight]{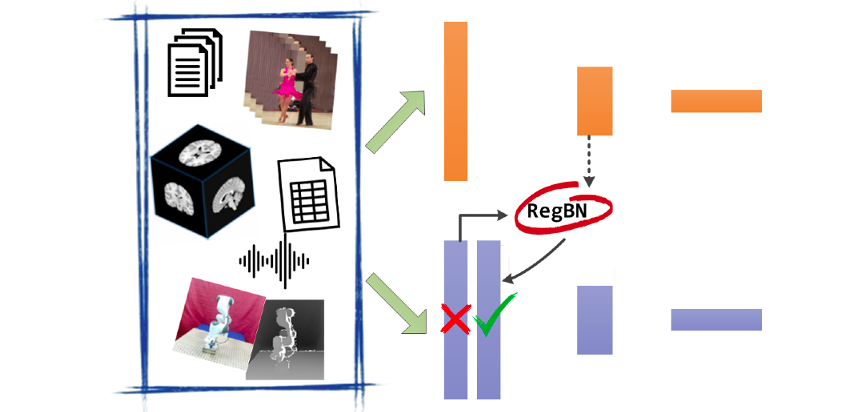} 
         \caption{RegBN as a layer normalizer in a multimodal model}
         \label{fig:appendix-RegBNlayer}
     \end{subfigure}
     \\
     \begin{subfigure}{1\columnwidth}
         \centering
         \includegraphics[height=0.2\textheight]{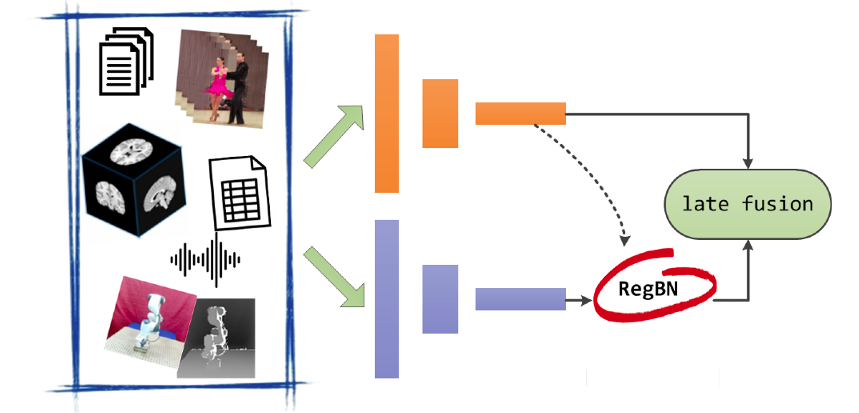}
         \caption{Late fusion with RegBN}
         \label{fig:appendix-fusionlate}
     \end{subfigure}
     \\
     \begin{subfigure}{1\columnwidth}
         \centering
         \includegraphics[height=0.2\textheight]{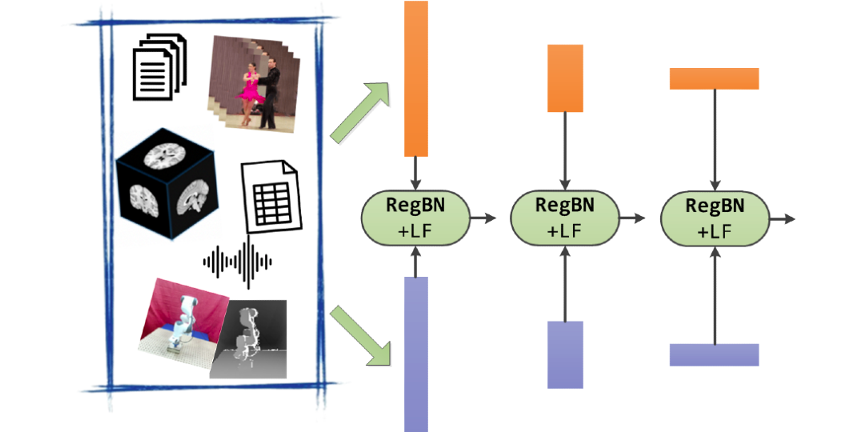}
         \caption{Layer fusion (LF) with RegBN}
         \label{fig:appendix-fusionmid}
     \end{subfigure}
     \\
     \begin{subfigure}{1\columnwidth}
         \centering
         \includegraphics[height=0.2\textheight]{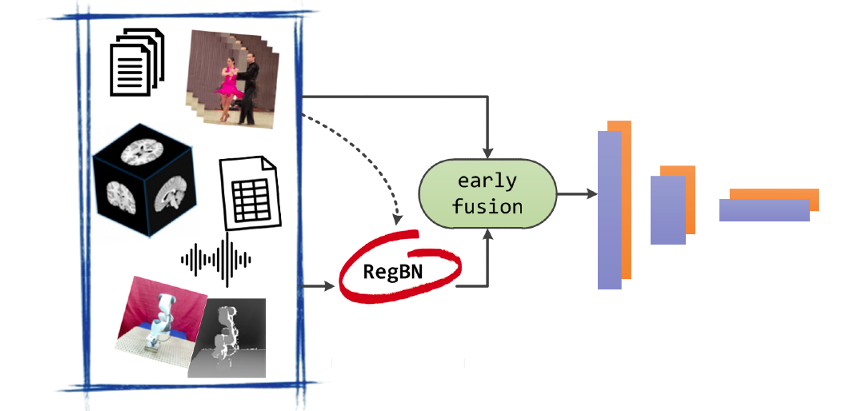}
         \caption{Early fusion with RegBN}
         \label{fig:appendix-fusionearly}
     \end{subfigure}
\end{center}
\caption{RegBN, as a multimodal normalization technique, in the context of different fusion paradigms. Given two modalities, $\mathcal{A}$ and $\mathcal{B}$, each modality has its respective pyramidal feature space (shown in different colors). (a) RegBN is capable of normalizing any pair of (input/hidden/output) layers in a multimodal neural network;  (b-d) a multimodal neural network fuses the features of modalities $\mathcal{A}$ and $\mathcal{B}$ in various ways, including early, layer, and late fusion. In each fusion scenario, the inputs to the fusion block are normalized using the RegBN technique. }
\label{fig:appendix-fusion}
\end{figure}

Multimodal data fusion can be accomplished at various feature levels, including low-level, high-level, or latent space. Furthermore, these features can be integrated through different fusion strategies, such as early, middle, or late fusion~\cite{wang2020deep,nagrani2021attention}. 
In multimodal research, determining the optimal fusion model and feature level for fusion remains challenging. 
This study showed that normalizing the features extracted from heterogeneous data sources can yield better fusion results. 
We demonstrate the applicability of RegBN as a multimodal normalization technique in various fusion structures within multimodal models. As depicted in Figure~\ref{fig:appendix-fusion}, we highlight several scenarios where RegBN can be employed effectively.

\begin{itemize}
    \item \emph{Layer normalization}: RegBN, as a normalization method, can be applied to any pair of multimodal layers regardless of their dimension and feature types, alleviating superimposed layers (caused by confounding factors and dependencies at both low- and high-level features) and thereby enhancing their efficiency (see Figure~\ref{fig:appendix-RegBNlayer}).  
    The experiments conducted on the synthetic dataset (Section~\ref{sec:synth}) and the healthcare diagnosis (Section~\ref{sec:diagnosis}) serve as illustrative instances of layer normalization.
    \item \emph{Late fusion}: 
This category of fusion is more popular among multimodal techniques. As shown in Figure~\ref{fig:appendix-fusionlate}, with the assistance of RegBN, the pair of output feature layers can be rendered independent, enabling the multimodal model to seamlessly combine the superimposed layers with enhanced efficiency. 
The utilization of RegBN in the late fusion structure can be observed in the domains of multimedia (Section~\ref{sec:multimedia}), affective computing (Section~\ref{sec:affective}), and robotics (Section~\ref{sec:robotics}).
    \item \emph{Layer fusion}: RegBN also facilitates layer-to-layer fusion. In this scenario, the corresponding layers are made mutually independent through RegBN before the fusion process takes place (see Figure~\ref{fig:appendix-fusionmid}). An example is provided with the synthetic experiment in Section~\ref{sec:synth}.
    \item \emph{Early fusion}: RegBN ensures the independence of input data before fusion in multimodal scenarios, leveraging the potential information of each source. As illustrated in Figure~\ref{fig:appendix-fusionearly}, the early fusion strategy is recommended when the input data exhibits high correlation, as demonstrated in the healthcare diagnosis experiment outlined in Section~\ref{sec:diagnosis}.
\end{itemize}

\section{Solution details}\label{appendix:sol}
To detail the solution, we first recall Eq.~\ref{eq:lambda}. The goal is to find the solution of the following objective:
\begin{equation}\label{eq:appendix-obj}
    \mathcal{F}(W^{(l,k)},\lambda_{+}) = \norm{f^{(l)}-W^{(l,k)} g^{(k)}}^2_2+\lambda_{+}
    \left(\norm{W^{(l,k)}}_F-1\right).
\end{equation}
This equation yields a closed-form solution for $W^{(l,k)}$:
\begin{equation}
\label{eq:appendix-omega}
    \hat{W}^{(l,k)} = \left(g^{(k)\top} g^{(k)}+\hat{\lambda}_{+}\mathbf{I}\right)^{-1}g^{(k)\top} f^{(l)},
\end{equation} 
where $\hat{\lambda}_{+}$ is obtained through the following:
\begin{equation}\label{eq:appendix-lambda}
    \hat{\lambda}_{+} = \underset{{\lambda}_{+}}{\mathrm{arg\,min}} \left(\norm{\bigl(g^{(k)\top} g^{(k)}+{\lambda}_{+}\mathbf{I}\bigr)^{-1}g^{(k)\top} f^{(l)}}_F-1\right).
\end{equation} 
One can use SVD for simplifying Eqs.~\ref{eq:appendix-omega} \& \ref{eq:appendix-lambda}. Let SVD decompose layer $g^{(k)}$:  SVD$\left(g^{(k)} \right) = U \Sigma V^* = \sum_{i=1}^m \sigma_{i} {u_i} {v^*_i}$, then the projection weights are re-written as follows
\begin{equation}\label{eq:appendix-omega-final}
    \hat{W}^{(l,k)}=\sum_{i=1}^{m}{\frac{\sigma_{i} }{\sigma_{i}^2+\hat{\lambda}_+}{u_i} {v_i} f^{(l)}},
\end{equation} 
where
\begin{equation}\label{eq:appendix-lambda2}
    \hat{\lambda}_{+} = \underset{{\lambda}_{+}}{\mathrm{arg\,min}} \left(\norm{
     \sum_{i=1}^{m}{\frac{\sigma_{i}}{\sigma_{i}^2+{\lambda}_+} {u_i} {v_i}f^{(l)}}}_F-1
     \right).
\end{equation} 
The L-BFGS algorithm facilitates the estimation of $\hat{\lambda}_{+}$ that is then inserted into Eq \ref{eq:appendix-omega-final} to compute the projection matrix. 
In our implementation, we employed the L-BFGS solver provided by PyTorch\footnote{The L-BFGS solver can be found at \url{https://pytorch.org/docs/stable/generated/torch.optim.LBFGS.html}.}. The default settings of the L-BFGS parameters are presented in Appendix~\ref{appendix:regbn}.

\section{Datasets}\label{appendix:datasets}
Below, we provide a brief overview of the databases used in this study. 
\subsection{LLP}\label{appendix:LLP}
``Look, Listen, and Parse'' (LLP)~\cite{tian2020unified} consists of 11,849 YouTube video clips, encompassing 25 event categories and totaling 32.9 hours of content sourced from AudioSet~\cite{gemmeke2017audio}. The LLP dataset includes 11,849 video-level event annotations, indicating the presence or absence of different video events to facilitate weakly-supervised learning. Each video has a duration of 10 seconds and contains at least 1 second of audio or visual events. Among the videos, 7,202 contain events from more than one event category, with an average of 1.64 different event categories per video. 
The samples were annotated for training with sparse labels, while the test set provides dense sound event labels for both video and audio at the frame level. 
 According to \cite{tian2020unified}, individual audio and visual events were annotated with second-wise temporal boundaries for a randomly selected subset of 1,849 videos from the LLP dataset to evaluate audio-visual scene parsing performance. The audio-visual event labels were derived from the audio and visual event labels. There are 6,626 event annotations, including 4,131 audio events and 2,495 visual events, for the 1,849 videos. Merging the individual audio and visual labels results in 2,488 audio-visual event annotations. This subset is divided into 649 videos and 1,200 videos for validation and inference, respectively. 
The AVVP~\cite{tian2020unified} framework was trained on the 10,000 videos with weak labels, and the trained models were tested on the validation and testing sets with fully annotated labels. 
The audio and visual features of videos in the LLP dataset are openly accessible\footnote{\url{https://github.com/YapengTian/AVVP-ECCV20}}.

\subsection{MM-IMDb}\label{appendix:MM-IMDb}
The Multimodal IMDb (MM-IMDb)~\cite{arevalo2017gated} dataset comprises two modalities, namely text and image. The aim is to perform multi-label classification on this dataset and predict the movie genre using either an image or text modality. This task is challenging as a movie can be assigned multiple genres. The dataset consists of 25,956 movies classified into 23 different genres. We use the same training and validation splits as in the previous study by \cite{ma2021smil}. 
The RGB images underwent standardization by rescaling from $261\times 385$ to $256\times 256$. They were then cropped to achieve a size of $224\times 224$. 
The dataset utilized in this study is openly accessible and available for public use\footnote{\url{https://github.com/pliang279/MultiBench}}.
\subsection{AV-MNIST}\label{appendix:AV-MNIST}
AV-MNIST\footnote{The dataset is available at \url{https://github.com/pliang279/MultiBench}.} is generated by combining spoken digit audio from the Free Spoken Digits Dataset\footnote{\url{https://github.com/Jakobovski/free-spoken-digit-dataset}} with written digits from the MNIST dataset\footnote{\url{http://www.pymvpa.org/datadb/mnist.html}}.
The objective is to classify the digit into one of ten categories (0 - 9). Classification of this dataset is a challenging task as the visual modality's energy is reduced by 75\%  through PCA and real-world background noises are added into the audio modality. 
The grey images in AV-MNIST have a resolution of $28\times28$ pixels, while the audio spectrogram is $112\times112$ pixels. 
\subsection{IEMOCAP}\label{appendix:IEMOCAP}
The ``Interactive Emotional Dyadic Motion Capture'' (IEMOCAP) dataset~\cite{busso2008iemocap} comprises 151 videos focused on dyadic interactions for human emotion analysis. 
The dataset comprises around 12 hours of audiovisual data, encompassing various modalities such as video, speech, motion capture of the face, and text transcriptions. 
The IEMOCAP database is annotated by multiple annotators, assigning categorical labels to the data, including emotions such as anger, happiness, sadness, and neutrality, and dimensional labels such as valence, activation, and dominance. 
In line with study \cite{tsai2019multimodal}, four emotions were chosen, including happy, sad, angry, and neutral, which are used for emotion recognition. It is important to note that IEMOCAP is a multilabel task, meaning that multiple emotions can be assigned to an individual. 
The dataset's multimodal streams have fixed sampling rates for audio signals at 12.5 Hz and for vision signals at 15 Hz. In line with study~\cite{tsai2019multimodal}, the evaluation of the dataset involves reporting the binary classification accuracy and the F1 score of the predictions.

\subsection{CMU-MOSI}\label{appendix:MOSI}
The Multimodal Opinionlevel Sentiment Intensity (CMU-MOSI)~\cite{zadeh2016mosi} is a comprehensive compilation of 2,199 short monologue video clips, each meticulously labeled with various annotations. 
These annotations include subjectivity labels, sentiment intensity labels, per-frame visual features, per-opinion visual features, and per-millisecond audio features. 
The CMU-MOSI dataset serves as a realistic and practical multimodal dataset for the task of affect recognition. It is widely utilized in competitions and workshops focusing on affect recognition research. The preprocessed versions of the CMU-MOSI dataset are openly accessible\footnote{\url{https://github.com/pliang279/MultiBench}}. 
The annotation of sentiment intensity within the dataset encompasses a comprehensive range from -3 to +3. This extensive annotation range enables the development of fine-grained sentiment prediction capabilities that go beyond the conventional positive/negative categorization, facilitating a more nuanced understanding of sentiment in the data. 
The videos within the dataset feature a diverse group of 89 speakers, with a distribution of 41 female speakers and 48 male speakers. These speakers represent a range of backgrounds, including Caucasian, African-American, Hispanic, and Asian individuals. Most speakers fall within the age range of 20 to 30 years. It is worth noting that all speakers in the dataset express themselves in English, and the videos originate from either the United States of America or the United Kingdom. 
Age, gender, and race are confounding factors. 
The training set consists of 52 videos, the validation set contains 10 videos, and the test set comprises 31 videos. Splitting these videos into segments results in a total of 1,284 segments in the training set, 229 segments in the validation set, and 686 segments in the test set. 

\subsection{CMU-MOSEI}\label{appendix:MOSEI}
CMU Multimodal Opinion Sentiment and Emotion Intensity (CMU-MOSEI)\footnote{\url{https://github.com/A2Zadeh/CMU-MultimodalDataSDK}} is a large-scale dataset specifically designed for sentence-level sentiment analysis and emotion recognition in online videos, comprising over 65 hours of annotated video data sourced from a diverse set of over 1,000 speakers and covering more than 250 topics. 
Videos in the CMU-MOSEI dataset were meticulously annotated for sentiment, along with identifying nine distinct emotions: anger, excitement, fear, sadness, surprise, frustration, happiness, disappointment, and neutrality. Continuous emotions such as valence, arousal, and dominance were also annotated. The inclusion of diverse prediction tasks makes CMU-MOSEI a highly valuable dataset for evaluating multimodal models across various affective computing tasks encountered in real-world scenarios. 
This experiment divided the dataset into three subsets: training, validation, and test sets. The training set contained 16,265 samples, the validation set had 1,869 samples, and the test set comprised 4,643 samples. 
The dimensions of the text data, audio data, and vision data are as follows: the text data has dimensions of $50\times300$, the audio data has dimensions of $500\times74$, and the vision data has dimensions of $500\times35$. 
The accuracy of the results with the CMU-MOSEI and CMU-MOSI datasets is assessed using the following evaluation metrics:
\begin{itemize}
\item Binary accuracy (Acc$_2$): This metric indicates the accuracy of sentiment classification as either positive or negative.
\item  5-Class accuracy (Acc$_5$): In this evaluation, each sample is labeled by human annotators with a sentiment score ranging from -2 (indicating strongly negative) to 2 (representing strongly positive). 
\item  7-Class accuracy (Acc$_7$): Similar to Acc$_5$, each sample is labeled with a sentiment score ranging from -3 (strongly negative) to 3 (strongly positive). 
\end{itemize}

\subsection{ADNI}\label{appendix:ADNI}
The ``Alzheimer’s Disease Neuroimaging Initiative'' (ADNI)\footnote{\url{https://adni.loni.usc.edu}} database is the most popular benchmark for Alzheimer's research and diagnosis. 
The dataset includes both 3D MRI scans and tabular metadata. The objective is to categorize patients into three groups: cognitively normal (CN), mildly cognitively impaired (MCI), or Alzheimer's demented (AD). 
The dataset was prepared in line with ~\cite{wolf2022daft}. 
T1-weighted MRIs were first normalized with minimal pre-processing and then segmented using the FreeSurfer v5.3 software\footnote{\url{https://surfer.nmr.mgh.harvard.edu/fswiki/DownloadAndInstall5.3}}. Only the regions of size $64\times 64\times 64$ around the left hippocampus were extracted since this region is strongly affected by Alzheimer’s disease. 
The tabular data as metadata comprises nine variables that contribute valuable information to the dataset. These variables include ApoE4, which indicates the presence or absence of the Apolipoprotein E4 allele known to be associated with an elevated risk of Alzheimer's disease. The dataset also contains variables related to cerebrospinal fluid biomarkers, P-tau181 and T-tau, which provide insights into the pathological changes associated with the disease. Additionally, demographic variables such as age, gender, and education are included. Two derived measures, obtained from 18F-fluorodeoxyglucose (FDG) and florbetapir (AV45) PET scans, serve as summary measures in the dataset. 
In accordance with the methodology presented in the work by~\citet{wolf2022daft}, the dataset underwent a partitioning process into five distinct folds. This partitioning strategy was carefully devised to ensure a balanced distribution of diagnosis, age, and sex across the folds. 
During the evaluation process, each of the five folds was used as a test set once, ensuring comprehensive coverage across the dataset. The remaining folds were further divided into five equally balanced chunks. From these chunks, one chunk was randomly selected to serve as the validation set, while the remaining data within the folds constituted the training set. As a result, the resulting distribution of data splits consists of 20\% for the test set, 16\% for the validation set, and 64\% for the training set. This partitioning scheme aims to maintain a proportional and representative distribution of the data subsets, facilitating reliable evaluation and training procedures.

\subsection{Vision\&Touch}\label{appendix:Vision&Touch}
Vision\&Touch, introduced by ~\citet{lee2019making}, is a collection of real-world robot manipulation data. It encompasses visual, force, and robot proprioception information. The data is obtained by executing two policies on the robot: a random policy that takes random actions and a heuristic policy that aims to perform peg insertion. The dataset includes several sensor modalities: robot proprioception, RGB-D camera images, and force-torque sensor readings. The proprioceptive input comprises the pose of the robot's end-effector, as well as linear and angular velocity. 
RGB images and depth maps are captured using a fixed camera, which is positioned to focus on the robot.  
The force sensor provides feedback on six axes, measuring the forces and moments along the x, y, and z axes. 
The primary objective of the dataset was to support representation learning specifically for reinforcement learning applications. 
The Vision\&Touch dataset was split into training, validation, and testing with 36,499, 4,047, and 22,800 samples, respectively. 
The dimensions of the RGB and depth images are $128\times128\times3$ and $128\times128\times1$, respectively. Data of size $6\times32$ and $8$ is acquired from force-torque sensors and robot proprioception, respectively. 
The database is publicly accessible\footnote{\url{https://sites.google.com/view/visionandtouch}}.
\subsection{Synthetic dataset}\label{appendix:synth}
Inspired by~\cite{lu2021metadata}, we designed this dataset to evaluate a confounding variable's influence on the training and inference procedures of multimodal models. 
The primary objective of the synthetic dataset is a binary classification of two distinct groups of data, called Group 1 and Group 2. 
Each group consists of a collection of 5,000 images, each with a resolution of $64\times 64$ pixels. Each image was partitioned into four quadratic sub-images (i.e., $32\times 32$) by dividing the input image horizontally and vertically into two equal halves. All the sub-images were generated using a 2D Gaussian distribution with a standard deviation of 5. 
The magnitudes of sub-images in the main diagonal were multiplied by $\sigma_{cls}$, corresponding to the classification label. Similarly, the magnitudes of the bottom-left quadrant sub-image were multiplied by $\sigma_{c}$, which plays the role of a confounding factor. In line with the methodology proposed in~\cite{lu2021metadata}, the magnitudes of the top-right sub-image were multiplied by zero to simplify the experiment. 
The values of $\sigma_{cls}$ for images in Group 1 were randomly sampled from a uniform distribution $\mathcal{U}(1,5)$. Likewise, for images in Group 2, $\sigma_{cls}$ values were randomly sampled from a uniform distribution $\mathcal{U}(4,8)$. Due to the overlapping of labels between the two groups, the maximum achievable accuracy, in theory, is 87.5\%. 
Regarding the confounding variable $\sigma_{c}$, it was assigned a random number within the same range as the true label, i.e., $\sigma_{c}\in \mathcal{U}(1,5)$ for Group 1 and $\sigma_{c}\in \mathcal{U}(4,8)$ for Group 2. 
In addition to the images, metadata of length 16 was created. This metadata comprises the (true) binary label, the confounding variable  $\sigma_{c}$ value, one randomly generated fake binary label, twelve randomly generated floating-point fake features, and one column with a constant value of one. In the ideal scenario, a multimodal model should primarily consider the first column of the metadata for classification since it uses a loss cross-entropy function with regard to the true labels during training, and the values in the other columns do not provide significant information, just used as metadata.
Furthermore, the experiment was repeated for other values, referred to as Experiment II, where $\sigma_{cls}$ and $\sigma_{c}$ were uniformly sampled from the range $\mathcal{U}(1,7)$ for Group 1 and $\mathcal{U}(4,10)$ for Group 2. In Experiment II, the theoretical maximum accuracy is 75.0\%.

\section{Baseline methods with RegBN as multimodal normalization method}\label{appendix:methods}
Here, we review the baseline methods that are employed in this study while also detailing the application of the proposed normalization method in the normalization of the multimodal data.

\subsection{AVVP}\label{appendix:AVVP}

\begin{figure}
  \centering
        \includegraphics[width=0.85\textwidth]{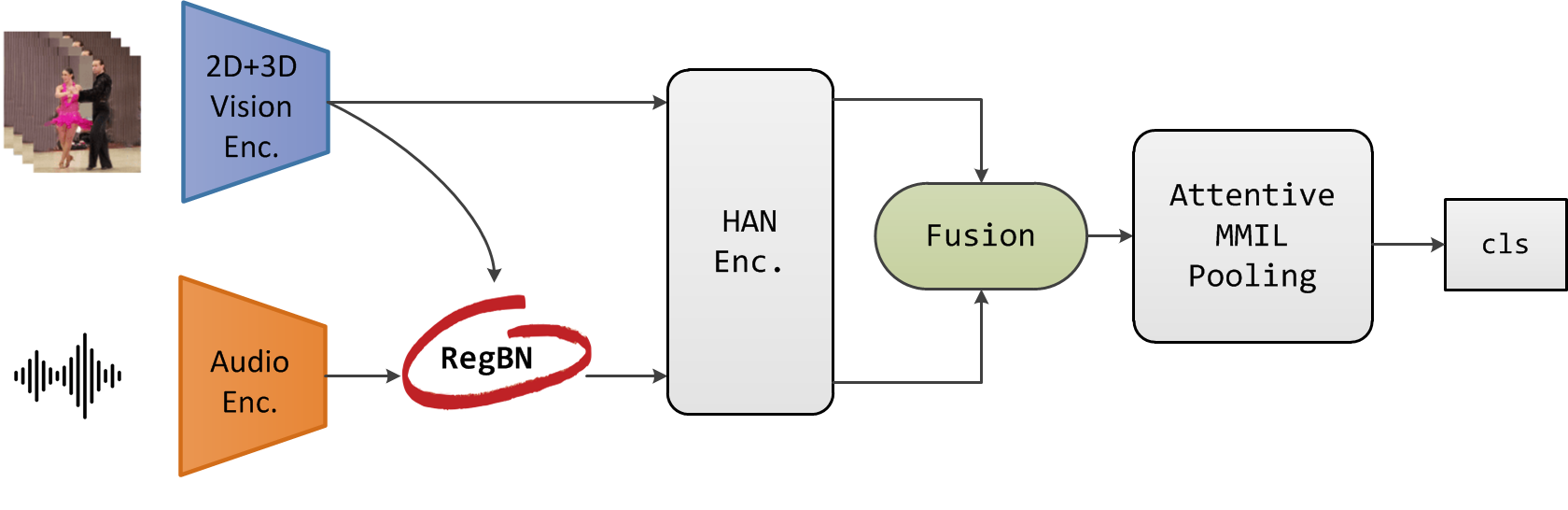}
\caption{The inference framework of AVVP~\cite{tian2020unified} with RegBN as a multimodal normalization method}
\label{fig:appendix-avvp}
\end{figure}

\citet{tian2020unified} proposed Audio-Visual Video Parsing (AVVP) for parsing individual audio, visual, and audio-visual events. 
The framework of AVVP is illustrated in Figure~\ref{fig:appendix-avvp}.  
Visual and snippet-level features  are extracted  from 2D frame-level and 3D snippet-level features using ResNet152\footnote{\url{https://pytorch.org/vision/main/models/generated/torchvision.models.resnet152.html}} and ResNet (2+1)D\footnote{\url{https://pytorch.org/vision/main/models/generated/torchvision.models.video.r2plus1d_18.html}}, respectively. 
Likewise, audio features are extracted via VGGish\footnote{\url{https://github.com/tensorflow/models/tree/master/research/audioset/vggish}}.
Audio, visual, and snippet-level feature dimensions are $10\times128$, $80\times2048$, and $10\times512$, respectively. 
The normalization module is employed to decouple the audio features from the visual ones, ensuring their independence. 
The AVVP method employs a hybrid attention network (HAN) to predict both audio and visual event labels based on the aggregated features. 
The HAN block incorporates both a self-attention network and a cross-attention network, enabling the HAN to dynamically learn which bimodal and cross-modal snippets to look for in every audio or visual snippet. 
An attentive multimodal multiple-instance learning (MMIL) pooling technique is employed to facilitate adaptive prediction of video-level event labels in weakly-supervised learning. This approach allows for adaptively pooling information from multiple instances in a multimodal manner. Additionally, an individual-guided learning strategy addresses the modality bias problem, ensuring fair representation and consideration of each modality in the learning process. 
For the final classification step, a fully-connected layer is utilized, followed by a sigmoid activation layer.

\subsection{SMIL}\label{appendix:SMIL}

\begin{figure}
  \centering
        \includegraphics[width=0.8\textwidth]{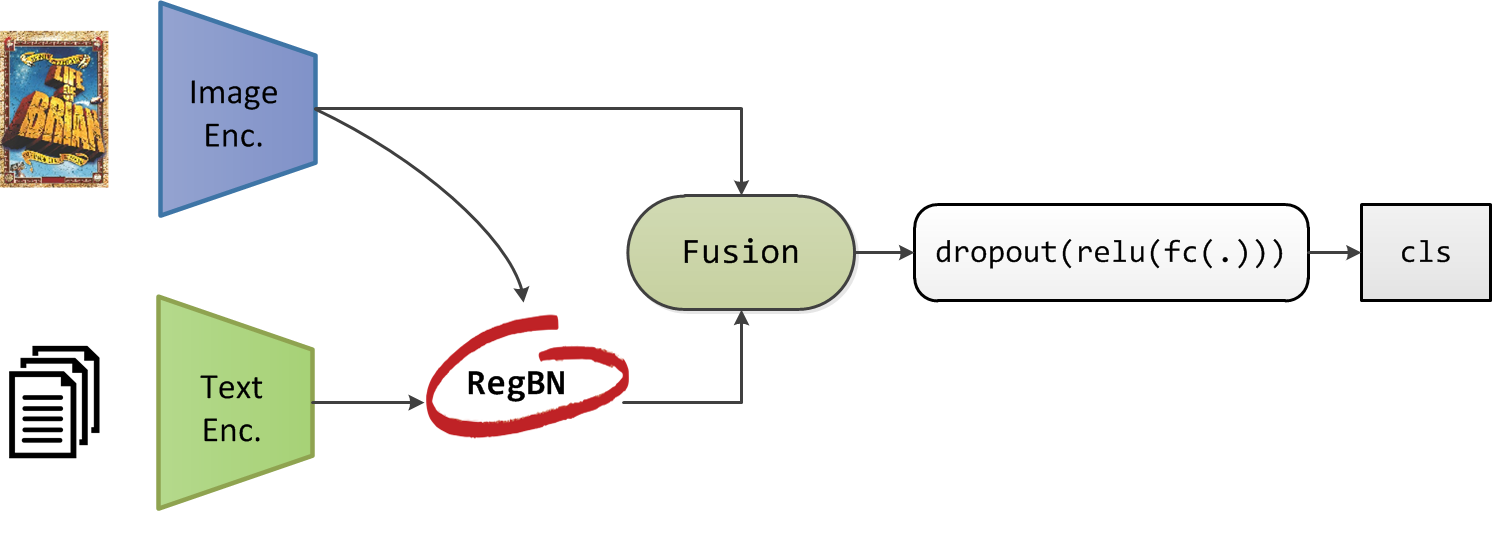}
\caption{The inference framework of SMIL~\cite{ma2021smil} with RegBN as a multimodal normalization method on MM-IMDb}
\label{fig:appendix-slimMMIMDb}
\end{figure}

\begin{figure}
  \centering
        \includegraphics[width=0.75\textwidth]{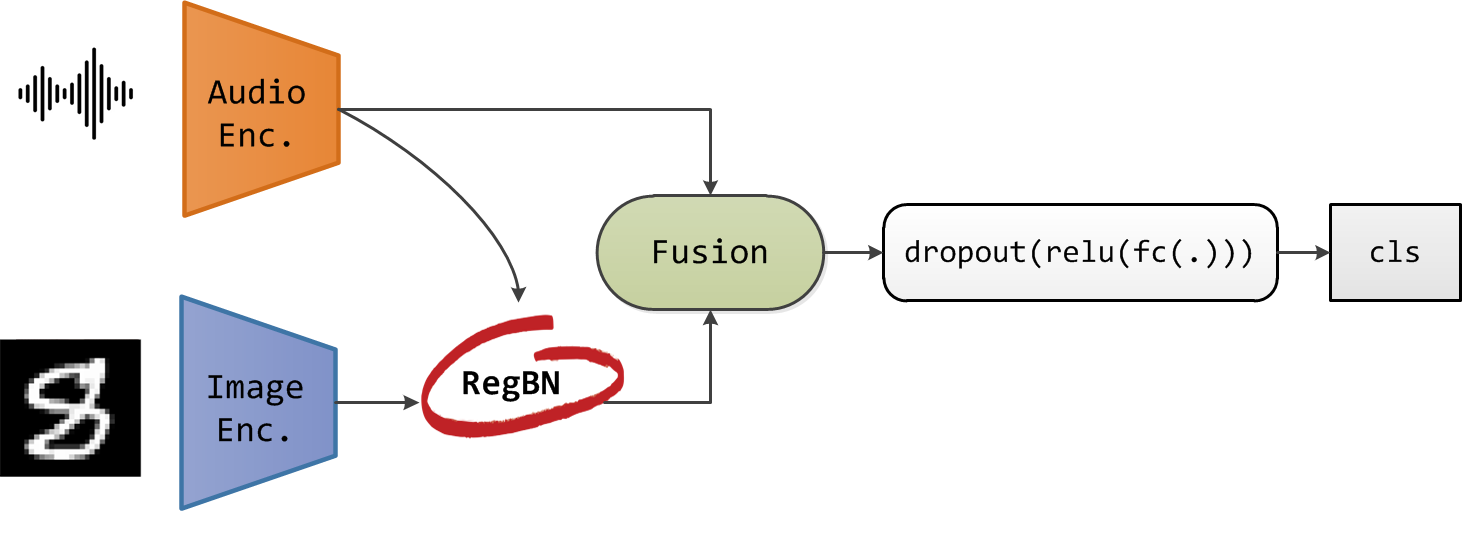}
\caption{The inference framework of SMIL~\cite{ma2021smil} with RegBN as a multimodal normalization method on AV-MNIST}
\label{fig:appendix-slimAV-MNIST}
\end{figure}

SMIL~\cite{ma2021smil}\footnote{\url{https://github.com/mengmenm/SMIL}} stands for multimodal learning with severely missing modality, a method for learning a multimodal model from a complete or an incomplete dataset. 
The framework of SMIL is illustrated in Figures~\ref{fig:appendix-slimMMIMDb}\&\ref{fig:appendix-slimAV-MNIST}. 
For the MM-IMDb dataset (Figure~\ref{fig:appendix-slimMMIMDb}), the textual data is converted to lowercase, followed by feature extraction using the pre-trained BERT\footnote{\url{https://github.com/google-research/bert}}$^,$\footnote{\url{https://huggingface.co/docs/transformers/index}} models. The length of text features is 768. 
Features of length 512  were extracted from the images utilizing the pre-trained VGG-19\footnote{\url{https://pytorch.org/vision/main/models/generated/torchvision.models.vgg19.html}} model. 
The text features are subsequently normalized with respect to the visual features using the proposed RegBN before the fusion process occurs. The fusion operation involves a concatenation layer. The fused features are passed through two fully-connected layers to obtain the classification labels. 
Likewise, a similar framework is employed for the audio-vision classification of the AV-MNIST dataset. 
As shown in Figure~\ref{fig:appendix-slimAV-MNIST}, modified LeNet-5 (refer to \cite{ma2021smil} for details) and LeNet-5 models are employed for extracting features, audio, and images, respectively. The feature dimension is 192 for audio features and 48 for vision features. 
We normalize the vision features with regard to the audio ones. Subsequently, the normalized vision features are concatenated with audio features and then fed into two fully-connected layers for classification.

\begin{figure}
  \centering
        \includegraphics[width=1.\textwidth]{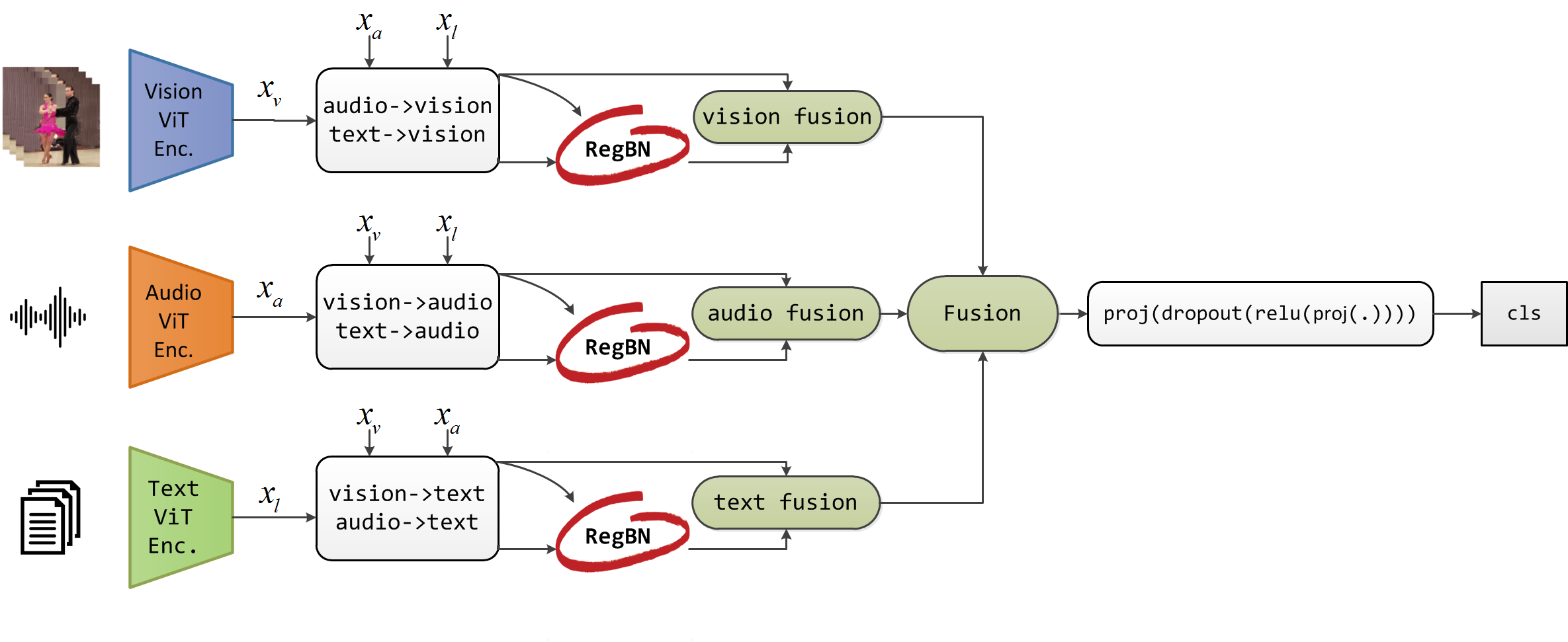}
\caption{The inference framework of MulT~\cite{tsai2019multimodal} with RegBN as a multimodal normalization method}
\label{fig:appendix-mult}
\end{figure}

\begin{figure}
  \centering
        \includegraphics[width=0.85\textwidth]{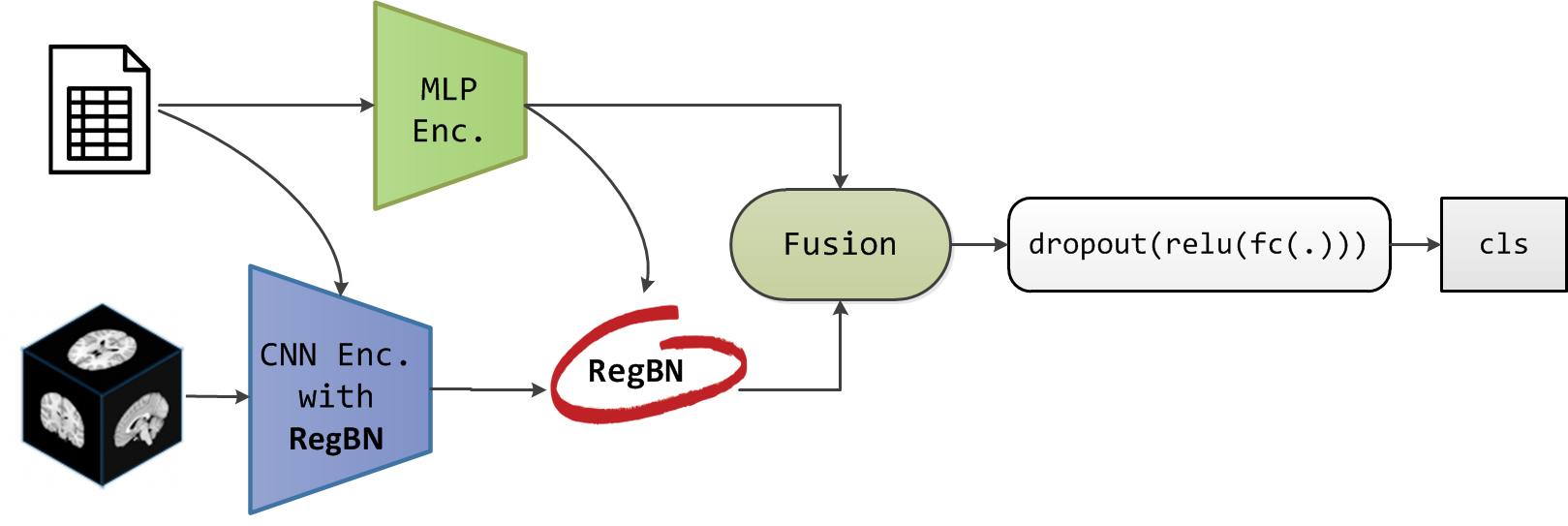}
\caption{The framework of healthcare diagnosis model with RegBN as a multimodal normalization method}
\label{fig:appendix-health}
\end{figure}

\subsection{MulT}\label{appendix:MulT}
MulT, introduced by \citet{tsai2019multimodal} in their work on multimodal language sequences, is a transformer-based model designed specifically for multimodal data representation. 
As shown in the framework of MulT in Figure~\ref{fig:appendix-mult},
this technique combines multimodal time series, including language/text ($l$), video/vision ($v$), and audio ($a$) modalities, through a feed-forward fusion mechanism using multiple directional pairwise crossmodal transformers. 
MulT has been developed to address the complexities associated with multimodal language sequences, which often exhibit an unaligned nature and require the inference of long-term dependencies across modalities. MulT is designed to handle both word-aligned and unaligned versions of these sequences. 
At a high-level feature, MulT effectively merges and integrates information from different modalities to capture the complex relationships within the multimodal data. 
Each crossmodal transformer in MulT serves to reinforce a target modality with low-level features from another modality by learning their attention-based interactions. 
This is done in three levels, including 1) language fusion ($v\rightarrow l$ \& $a\rightarrow l$), 2) audio fusion ($v\rightarrow a$ \& $l\rightarrow a$), and 3) video fusion ($a\rightarrow v$ \& $l\rightarrow v$). 
The MulT architecture encompasses crossmodal transformers that model interactions between all pairs of modalities. This is followed by sequence models, such as self-attention transformers, which use fused features for prediction. 
Before each level of fusion, RegBN is performed on the pair of multimodal with high-level features as a normalization block.

\subsection{Healthcare model}\label{appendix:MMCL-TI}
For the healthcare experiment, which is the diagnosis of Azhimere on the ADNI dataset, we utilize the 3D ResNet and MLP network proposed in studies by \citet{wolf2022daft} and \citet{hager2023best}, respectively. These networks extract 3D MRI and tabular features, respectively. 
As shown in Figure~\ref{fig:appendix-health}, we apply normalization techniques to the abovementioned models in low- and high-level features. 
The 3D ResNet consists of an input convolution layer and four ResNet blocks. During feature extraction from MRI data, the features undergo normalization using a technique listed in Table~\ref{table:appendix-dem}. The output feature vector from the 3D ResNet is of length 256. 
On the other hand, the MLP network applies a fully-connected layer, followed by a ReLU activation layer, three times to process the tabular data. The output feature vector from the MLP network is of length 16. Once again, the visual and tabular data are normalized at the high-level feature stage. This is followed by a combination of dropout, ReLU activation, and fully-connected layers before being fed into a classification block.

\begin{figure}
  \centering
        \includegraphics[width=0.75\textwidth]{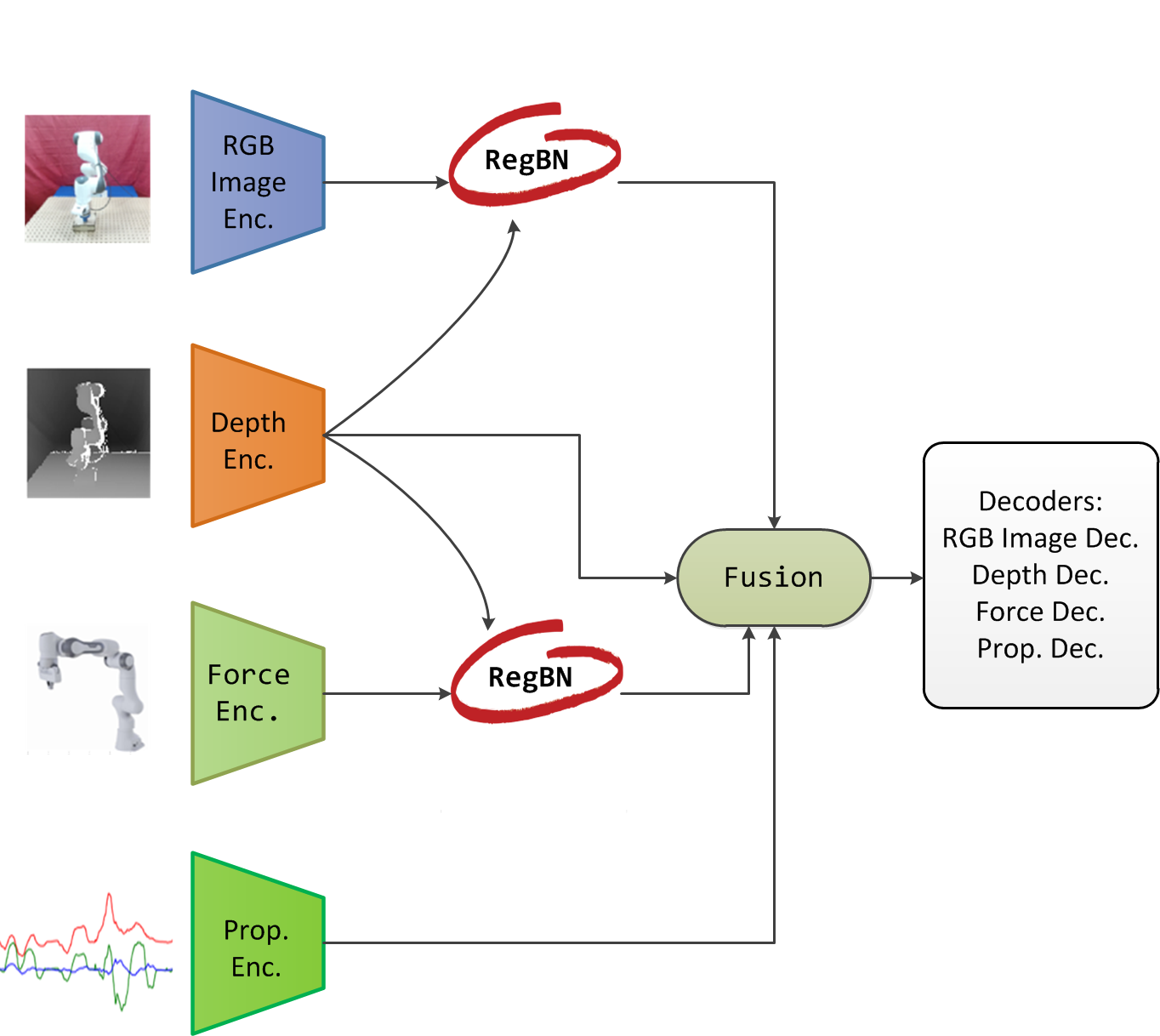}
\caption{The framework of the MSVT model with RegBN as a multimodal normalization method}
\label{fig:appendix-MSVT}
\end{figure}

\subsection{MSVT}\label{appendix:MSVT}
Making Sense of Vision and Touch (MSVT)~\cite{lee2019making} is a self-supervised multimodal approach aimed at acquiring a  representation of sensory multimodal inputs. This approach involves training an end-to-end representation learning network through self-supervision. As shown in Figure~\ref{fig:appendix-MSVT}, the MSVT method utilizes data from four distinct sensors: RGB images, depth maps, force-torque readings within a 32ms timeframe, and the position, orientation, and velocity of the robot's end-effector. These data inputs are encoded and fused into a multimodal representation using a variational Bayesian technique, enabling the learning of a policy for manipulation tasks involving contact-rich environments.

For the RGB images, a neural network with a convolutional neural network (CNN) backbone is employed. The CNN architecture comprises six convolutional layers with downsampling ($3\rightarrow16\rightarrow32\rightarrow64\rightarrow64\rightarrow128\rightarrow128$). Subsequently, the extracted features are flattened and passed through a fully-connected layer, resulting in a feature vector of length 256. 
Similarly, the depth data is processed using six convolutional layers with downsampling and one fully-connected layer. The output dimension of the depth features is 256. 
To account for the correlation between visual and depth data, the RGB image is subjected to normalization with respect to the depth image using the RegBNat technique at higher levels of feature representation. 
The force encoder module comprises five convolutional units, each consisting of a 1D convolution layer with downsampling followed by a LeakyReLU activation function with a negative slope of 0.1 ($6\rightarrow16\rightarrow32\rightarrow64\rightarrow128\rightarrow256$). The force encoder maps input data of size $32\times6$ to a feature vector of 256, capturing the sliding contact dynamics along the x, y, and z axes, which are superimposed on the depth information. To address this overlap, the force sensor data is normalized with respect to the depth features. 
The proprioception encoder module is composed of four blocks, with each block containing a fully-connected layer followed by a LeakyReLU activation function with a negative slope of 0.1 ($8\rightarrow32\rightarrow64\rightarrow128\rightarrow256$). 
The resultant of all the encoders are concatenated/fused. The reconstruction decoders are similar to the encoders.  More details can be found in \cite{lee2019making}\footnote{The code of MSVT is available at \url{https://github.com/stanford-iprl-lab/multimodal_representation}.}. 

\subsection{CNN and MLP networks for the synthetic experiment}\label{appendix:CNNandMLP}
Due to the simple structure of the synthetic data,  MLP and CNN models developed for this experiment are relatively light. 
The MLP model incorporates three consecutive convolution layers, without normalization, to process input $64\times64$ synthetic images. These convolution layers are followed by a fully-connected layer, which converts the features into a feature vector of length 128. Subsequently, the resultant vector is normalized with respect to the input tabular data using one of the techniques mentioned in Table~\ref{table:appendix-synth}. 
Consequently, the normalized feature vector is passed through a fully-connected layer with a sigmoid activation function for binary classification. 
The CNN architecture is structured as follows for processing input synthetic images:
Four convolutional layers are applied, each with a stride of 2. Each convolutional layer is equipped with a normalization technique mentioned in Table~\ref{table:appendix-synth}, followed by a ReLU activation function. 
The output from the convolutional layers is then flattened to convert it into a one-dimensional feature vector. This feature vector is then passed through a fully-connected layer with a ReLU activation function.
Finally, another fully-connected layer with a sigmoid activation function is applied for binary classification. 

\section{Experimental Results and Details}\label{appendix:results}
This section presents and provides a comprehensive account of the results that were not included in the paper due to page limit constraints. 
It is worth noting that we used the default settings of different normalization methods recommended. 
The batch size for most methods, including RegBN, was set to 50 in all the experiments conducted. However, it should be noted that the batch size for MDN was set to 200, which differs from the other methods. 

\subsection{RegBN parameter setting}\label{appendix:regbn}
To update its learnable projection matrix, RegBN employed the exponential moving average method and the L-BFGS optimizer that incorporates specific parameters (see Sections~\ref{sec:updateW}\&\ref{sec:avoidextrema}). Table~\ref{table:appendix-regbnparams} presents the predetermined parameters along with their respective constant values, which remained unchanged throughout the entirety of the experiment conducted in this study. 
In Section~\ref{appendix:abalation}, we present an ablation study on the predetermined parameters of RegBN.

\begin{table}
  \caption{The default setting for RegBN's parameters}
  \label{table:appendix-regbnparams}
  \centering
  \begin{tabular}{l c}
    \toprule
    parameter  & value\\
    \midrule
    \textbf{Parameters of exponential moving average}\\
    decay rate $\beta_1$ in Eq.~\ref{eq:w2} & 0.9\\
    decay rate $\beta_2$ in Eq.~\ref{eq:w2} & 0.99\\
    $\Lambda_p$  in Eq.~\ref{eq:lambdai}   & [1, 100, 1000]\\~\\
    \textbf{Parameters of the L-BFGS solver}\\
    learning rate& 1.0\\
    maximal number of iterations per optimization step & 25\\
    termination tolerance on first-order optimality &  0.00001\\
    \bottomrule
  \end{tabular}
\end{table}

\subsection{Multimedia}\label{appendix:multimedia}
The validation results of AVVP with the LLP dataset are presented in Table~\ref{table:appendix-multimedia-AVVP-train}. We report the validation results to show how AVVP can be efficiently trained in the presence of RegBN. 
RegBN improves the segment-level and event-level results of AVVP for audio (A), vision (V), and audio-vision (A-V) inputs, enhancing the performance across all modalities. 
Table~\ref{table:appendix-avmnist} showcases the classification accuracy achieved by SMIL on the AV-MNIST dataset. 
The table clearly demonstrates that incorporating multiple modalities leads to superior results compared to using individual modalities alone. By utilizing RegBN as a multimodal normalization technique, the performance of the baseline SMIL model is improved. This highlights the importance of normalizing multimodal data prior to fusion, emphasizing the necessity of such normalization for gaining better results.

\begin{table}
  \caption{
  Audio-visual video parsing's \emph{validation} accuracy (\%) of baseline AVVP~\cite{tian2020unified}, as baseline (BL), on the LLP dataset~\cite{tian2020unified} for different normalization techniques. AVVP has 4,571,723 learnable parameters. PMDN requires 26,214,400 parameters in its architecture, while RegBN does not need any learnable parameters as its learnable parameters are learned in a self-supervised way, described in Section~\ref{sec:proposed}.
}
  \label{table:appendix-multimedia-AVVP-train}
  \centering
  {\setlength{\tabcolsep}{4.5pt}
  \begin{tabular}{@{\extracolsep{4pt}}l ccccc ccccc@{}}
    \toprule
    Method  & \multicolumn{5}{c}{Segment-Level} &  \multicolumn{5}{c}{Event-Level} \\
    \cline{2-6} \cline{7-11} \addlinespace
    & A$\uparrow$ & V$\uparrow$  & A-V$\uparrow$  & Type$\uparrow$ & Event$\uparrow$ & A$\uparrow$ & V$\uparrow$  & A-V$\uparrow$  & Type$\uparrow$ & Event$\uparrow$\\
    \midrule
    BL       & 61.8 & 54.5 & 49 & {55.1} & 57.4 & {53.6} & 49.9 & 43.3 & {49.4} & 49.8\\
    BL+PMDN   & 61.4 & 54.6 & 48.9 & 54.8 & 57.4 & 52.8 & 50.5 & 43.3 & 49.3 & 50.2\\
    BL+RegBN  & {63.5} & {55.3} & {49.1} & 55 & {58} & 52.5 & {51.1} & {44} & 49 & {50.9}\\
    \bottomrule
  \end{tabular}
  }
\end{table}

\begin{table}
  \caption{Classification accuracy of baseline SMIL~\cite{ma2021smil} (denoted by BL) with/without normalization on the AV-MNIST dataset}
  \label{table:appendix-avmnist}
  \centering
  \begin{tabular}{l c c}
    \toprule
    Method  & Norm. params. ($\#$) & ACC(\%) $\uparrow$\\
    \midrule
    \textbf{Unimodal data}\\
    BL (Unimodal $i$) & - & 64.3\\
    BL (Unimodal $a$) & - & 43.2\\~
    \\
    \textbf{Multimodal data}\\
    BL  &  - &70.6\\
    BL+PMDN  & 11,136 & 70.7\\
    BL+RegBN  &  0 & 71.1\\
    \bottomrule
  \end{tabular}
\end{table}

\subsection{Affective computing}\label{appendix:affective}
Detailed results of MulT with and without RegBN for the IEMOCAP, CMU-MOSEI, and CMU-MOSI datasets can be found in Tables~\ref{table:appendix-Affective-IEMOCAP}, \ref{table:appendix-Affective-CMU-MOSEI}, and \ref{table:appendix-Affective-CMU-MOSI}, respectively. These tables comprehensively analyze the impact of incorporating RegBN in the MulT model. 
RegBN consistently improves the performance of MulT across various fusion categories. In both solo fusion and all fusion categories, RegBN enhances the results of MulT in most cases. Notably, significant improvements can be observed when using RegBN for audio or video fusion of word-aligned CMU-MOSI and for language fusion of IEMOCAP with word-aligned data. 
Furthermore, the inclusion of RegBN leads to substantial improvements in training and test loss values, demonstrating its effectiveness in enhancing the convergence of the multimodal model. The improved loss values indicate that RegBN plays a crucial role in optimizing the training process and ensuring better generalization performance during testing.

\begin{table}
  \caption{Detailed results of multimodal emotion analysis on IEMOCAP with \textbf{word-aligned} and \textbf{non-aligned} (inside parentheses) multimodal sequences. The baseline (BL) is MulT~\cite{tsai2019multimodal}.}
  \label{table:appendix-Affective-IEMOCAP}
{\setlength{\tabcolsep}{3pt}
  \centering
  \begin{tabular}{@{\extracolsep{4pt}}l cc cc cc cc cc@{}}
    \toprule
    Method  & \multicolumn{2}{c}{Loss} & \multicolumn{2}{c}{Happy (\%)} & \multicolumn{2}{c}{Sad (\%)} & \multicolumn{2}{c}{Angry (\%)} & \multicolumn{2}{c}{Natural (\%)} \\
    \cmidrule{2-3}  \cmidrule{4-5}  \cmidrule{6-7}  \cmidrule{8-9}  \cmidrule{10-11}  
    & Training$\downarrow$ & Test$\downarrow$ & Acc$\uparrow$ & F1$\uparrow$ & Acc$\uparrow$ & F1$\uparrow$ & Acc$\uparrow$ & F1$\uparrow$ & Acc$\uparrow$ & F1$\uparrow$\\
    \midrule
    \multicolumn{11}{c}{Language Fusion  ($v\rightarrow l$ \& $a\rightarrow l$)}\\
    BL &  0.141    &  0.458 & 85.6 & 84.1   & 83.1 & 81.8 & 83.8 & 83.1 & 67.5 & 66.6\\
     &   (0.381) & (0.63)  & (86.1) & (80.6)  & (79.9)  & (77.2) & (76.4)  & (70.2)  & (59.6) & (51.8) \\\addlinespace
    BL+RegBN  & 0.152 & 0.439  & 86.3 & 81.9  & 84.8  & 83.6 & 87.4 & 87.5 & 70.8 & 70.2\\
    &  (0.335) & (0.587)  & (86.1) & (80.6)  & (79.8)  & (76.9) & (76.0)  & (70.7)  & (60.0) & (55.6) \\
    %
    %
    \hdashline \addlinespace
    \multicolumn{11}{c}{Audio Fusion  ($v\rightarrow a$ \& $l\rightarrow a$)}\\
    BL   & 0.154 & 0.450  & 87.1 & 83.3  & 83.6  & 82.9 & 83.8 & 82.8 & 69.2 & 69.2\\
     & (0.394) & (0.678)  & (86.2) & (80.8)  & (79.6)  & (77.0) & (76.2)  & (69.9)  & (60.2) & (55.2) \\\addlinespace
    BL+RegBN  & 0.009 & 0.425  & 85.8 & 82.8  & 83.6  & 83.2 & 86.0 & 85.3 & 70.1 & 67.5\\
    &  (0.330) & (0.623)  & (86.4) & (80.6)  & (79.8)  & (76.9) & (76.3)  & (70.1)  & (59.7) & (54.5) \\
    %
    %
    \hdashline \addlinespace
    \multicolumn{11}{c}{Video Fusion  ($a\rightarrow v$ \& $l\rightarrow v$)}\\
    BL+Identity   & 0.141 & 0.498  & 87.2 & 86.1  & 83.9  & 83.9 & 85.0 & 84.8 & 70 & 69.7\\
     & (0.33) & (0.669)  & (86.1) & (80.6)  & (80.0)  & (76.9) & (76.4)  & (70.2)  & (59.6) & (56.2) \\\addlinespace
    BL+RegBN  & 0.189 & 0.477  & 86.6 & 84.5  & 81.2  & 79.3 & 83.1 & 82.5 & 68.7 & 68.4\\
    &   (0.285) & (0.645)  & (86.2)  & (80.5)  & (79.7) &(76.8)  &(76.3)  & (70)  & (59.8) & (54.7) \\
    \hdashline \addlinespace
    \multicolumn{11}{c}{All Fusion Categories}\\
    BL  & 0.106 & 0.536  & 86.3 & 84.0  & 81.5  & 80.6 & 86.5 & 86.4 & 69.5 & 69.1\\
     & (0.307) & (0.665)  & (85.1) & (80.4)  & (79.6)  & (77.0) & (76.4)  & (70.2)  & (59.8) & (55.6) \\\addlinespace
    BL+RegBN  & 0.009 & 0.452  & 87.4 & 83.0  & 84.3  & 84.1 & 88.2 & 88.1 & 73.4 & 73.2\\
    & (0.292) & (0.641)  & (86.1) & (80.6)  & (79.7)  & (76.8) & (76.3)  & (70.1)  & (59.8) & (54.9) \\
    \bottomrule
  \end{tabular} 
}
\end{table}

\begin{table}
  \caption{Multimodal sentiment analysis results multimodal on  CMU-MOSEI with \textbf{word aligned} and \textbf{non-aligned} (inside parentheses) multimodal sequences. The baseline (BL) method is MulT~\cite{tsai2019multimodal}.}
  \label{table:appendix-Affective-CMU-MOSEI}
  \centering
  \begin{tabular}{@{\extracolsep{4pt}}l cc ccccc@{}}
    \toprule
    Fusion   & \multicolumn{2}{c}{Loss} & \multicolumn{5}{c}{Sentiment (\%)} \\
    \cline{2-3}  \cline{4-8}
    & Training$\downarrow$  & Test$\downarrow$ & Acc$_2$$\uparrow$ & Acc$_5$$\uparrow$   & Acc$_7$$\uparrow$ & F1$\uparrow$  & Corr$\uparrow$\\
    \midrule
     \multicolumn{8}{c}{Language Fusion  ($v\rightarrow l$ \& $a\rightarrow l$)}\\
    BL   & 0.463 & 0.598 &  80.9  &  51.2 & 49.7 & 81.1  & 67.3\\
    &   (0.435) & (0.618) & (81.0) & (51.1) & (49.5) & (81.0)  & (67.2) \\\addlinespace
    BL+RegBN & 0.435 & 0.592  & 80.9 & 51.7  &  50.3 & 81.2  & 66.4\\
    &   (0.387) & (0.600) & (81.0) & (51.8) & (50.5) & (81.2)  & (67.7)\\
    %
    %
    \hdashline \addlinespace
    \multicolumn{8}{c}{Audio Fusion  ($v\rightarrow a$ \& $l\rightarrow a$)}\\
    BL   & 0.480 & 0.625  & 81.3 & 51.0  & 49.5 & 81.3  & 66.3\\
    &   (0.488) & (0.608) & (81.7) & (50.7) & (49.2) & (81.9)  & (67.7)\\\addlinespace
    BL+RegBN & 0.462 & 0.629  & 81.6 & 51.3  &  49.8 &  81.7  & 66.7\\
    &   (0.496) & (0.641) & (82.5) & (51.5) & (49.8) & (82.0)  & (66.5)\\
    %
    %
    \hdashline \addlinespace
    \multicolumn{8}{c}{Video Fusion  ($a\rightarrow v$ \& $l\rightarrow v$)}\\
    BL   & 0.451 & 0.632  & 80.8 & 52.1  &  50.7 & 81.0  & 65.8\\
    &   (0.48) & (0.617) & (80.6) & (50.2) & (48.9) & (80.9)  & (65.7) \\\addlinespace
    BL+RegBN & 0.418 & 0.626  & 80.7 & 51.2  & 49.8 & 81.4  & 67.0\\
    &   (0.465) & (0.615) &(81.4) & (51.9) & (50.4) & (81.7)  & (68.0)\\
    \hdashline \addlinespace
    \multicolumn{8}{c}{All Fusion Categories}\\
    BL   & 0.452 & 0.636 & 80.3 & 51.9 & 50.3  & 80.1  & 67.1\\
      & (0.481) & (0.619)  & (81)& (51.4) & (49.7)  & (81.2) & (67.5)\\\addlinespace
    BL+RegBN  & 0.438 & 0.611 & 81.1 & 52.2 & 50.5  & 81.24  & 66.6\\
    & (0.453) & (0.605) & (81.4) & (52.5) & (51.2)  & (81.6)  & (68.3)\\
    \bottomrule
  \end{tabular}
\end{table}

\begin{table}
  \caption{Multimodal sentiment analysis results multimodal on CMU-MOSI with \textbf{word aligned} and \textbf{non-aligned} (inside parentheses) multimodal sequences. The baseline (BL) method is MulT~\cite{tsai2019multimodal}.}
  \label{table:appendix-Affective-CMU-MOSI}
  \centering
  {\setlength{\tabcolsep}{4.5pt}
  \begin{tabular}{@{\extracolsep{4pt}}l cc ccccc@{}}
    \toprule
    Fusion   & \multicolumn{2}{c}{Loss} & \multicolumn{5}{c}{Sentiment (\%)} \\
    \cline{2-3}  \cline{4-8} \addlinespace
    & Training$\downarrow$  & Test$\downarrow$ & Acc$_2$$\uparrow$ & Acc$_5$$\uparrow$   & Acc$_7$$\uparrow$ & F1$\uparrow$  & Corr$\uparrow$\\
    \midrule
    \multicolumn{8}{c}{Language Fusion  ($v\rightarrow l$ \& $a\rightarrow l$)}\\
    BL   & 0.524 & 0.480  & 79.8 & 42.3  &  36.3  & 79.7  & 65.5\\
    &   (0.541 & (0.550) & (78.3) & (42.2) & (35.9) & (78.6)  & (59.4)\\\addlinespace
    BL+RegBN & 0.432 & 0.461  & 79.7 & 41.9  &  35.5 &  80.2  & 66.6\\
    &   (0.409) & (0.465) & (79.3) & (40.6) & (34.3) & (79.2)  & (66.3)\\
    %
    %
    \hdashline \addlinespace
    \multicolumn{8}{c}{Audio Fusion  ($v\rightarrow a$ \& $l\rightarrow a$)}\\
    BL   & 0.508 & 0.501  & 79.6 & 44.5  &  38.1 & 81.1  & 67.2\\
    &   (0.421) & (0.543) & (77.8) & (40.3) & (35.2) & (77.5)  & (63.1)\\ \addlinespace
    BL+RegBN & 0.376 & 0.531  & 80.7 & 51.3  &  49.8 & 81.4  & 66.6\\
    &   (0.329) & (0.495) & (80.2) & (40.9) & (35.4) & (80.5)  & (65.9)\\
    %
    %
    \hdashline \addlinespace
    \multicolumn{8}{c}{Video Fusion  ($a\rightarrow v$ \& $l\rightarrow v$)}\\
    BL   & 0.447 & 0.535  & 80.4 & 44.8  &  41.2 & 80.4  & 66.7\\
    &   (0.340) & (0.537) & (77.4) & (42.2) & (37.0) & (77.1)  & (63.4)\\ \addlinespace
    BL+RegBN & 0.403 & 0.524  & 80.7 & 51.2  &  49.8 & 81.4  & 67.0\\
    &   (0.321) & (0.525) & (79.1) & (40.6) & (36.4) & (78.9)  & (67.1)\\
    \hdashline \addlinespace
    \multicolumn{8}{c}{All Fusion Categories}\\
    BL   & 0.403 & 0.632 & 81.4 & 42.5  &  37.5 & 82.0  & 69.7\\
    &   (0.431) & (0.520) & (79.5) & (42.0) & (38.9) & (80.7)  & (67.3) \\\addlinespace
    BL+RegBN & 0.267 & 0.546 &81.8 & 42.3  &  38.6 & 82.3  & 69.1\\
    &   (0.401) & (0.481) & (81.5) & (42.9) & (39.6) & (82.0)  & (68.2)\\
    \bottomrule
  \end{tabular}
  }
\end{table}

\subsection{Healthcare diagnosis}\label{appendix:dem}
Table~\ref{table:appendix-dem} reports the performance of the baseline multimodal models developed in~\cite{hager2023best,wolf2022daft} with different normalization techniques. 
Healthcare data is most often accompanied by confounding effects, and Table~\ref{table:appendix-dem} suggests that conventional normalization methods such as BN, IN, and GN may not be suitable for effectively normalizing multimodal data. 
Multimodal data requires dedicated and specialized normalization methods tailored to their unique characteristics. 
Both MDN and PMDN demonstrate superior performance compared to conventional normalization techniques. However, the results are significantly improved when incorporating RegBN in terms of test accuracy and training loss. 
The RegBN results emphasize that it can effectively handle the complexities of multiple modalities.

\begin{table}
  \caption{Training cross-entropy (CE) loss,   test accuracy (ACC), and test balanced accuracy (BA) on the ADNI dataset~\cite{jack2008alzheimer}. Baseline is a combination of techniques developed in~\cite{hager2023best,wolf2022daft}. 
  }
  \label{table:appendix-dem}
  \centering
  \begin{tabular}{@{\extracolsep{4pt}}lcccc@{}}
    \toprule
     Method  &Norm. Params. & Training CE  & \multicolumn{2}{c}{Test}  \\
    \cmidrule{4-5} 
    & ($\#$) &loss$\downarrow$ & ACC (\%) $\uparrow$ & BA (\%)$\uparrow$ \\ 
    \midrule
    BL+BN    & 512 &  0.641±0.01  &  48.8±4.4 & 48.3±4.5\\
    BL+GN    & 512 &  0.649±0.01  &  48.4±4.6 & 47.9±4.5\\
    BL+LN    & 512 &  0.646±0.01  &  48.5±4.4 & 48.1±4.5\\
    BL+IN    & 512 &  0.647±0.01  &  48.2±4.5 & 48.0±4.4\\
    BL+MDN   & 0   &  0.619±0.03  & 50.4±4.8 & 50.1±5.6 \\
    BL+PMDN  & 4,096 & 0.632±0.03 & 49.7±6.2 & 49.6±7.5 \\
    BL+RegBN & 0 & {0.596±0.01} & {53.0±3.1} & {52.3±3.7} \\
    \bottomrule
  \end{tabular}
\end{table}

\subsection{Robotics}\label{appendix:robotics}
As mentioned in Appendix~\ref{appendix:MSVT} and depicted in Figure~\ref{fig:appendix-MSVT}, a pair of normalization units are employed for decoupling the RGB image-depth and force-depth pairs. 
These units are referred to as \emph{vision} and \emph{touch}, respectively, and the results are provided in Table~\ref{appendix:TouchVision}. 
The observed decrease in training loss values and simultaneous increase in test accuracy clearly indicate the necessity of the mentioned normalization units. 

\begin{table}
  \caption{Training and test results of MSVT~\cite{lee2019making}, as the baseline (BL) method, with different normalization methods on Vision\&Touch~\cite{lee2019making}. \emph{vision} and \emph{touch} refer to the normalization of RGB image-depth and
  force-depth pairs, respectively.}
  \label{appendix:TouchVision}
  \centering
  \begin{tabular}{@{\extracolsep{4pt}}l cccc@{}}
    \toprule
     Method  & Normalisation  & \multicolumn{2}{c}{Training} & Test Accuracy \\
    \cline{3-4} 
    & parameters (\#) & flow loss & total loss & (\%)  \\
    \midrule
    BL     & -- &  0.212 & 0.563 &  86.22\\
    BL+PMDN (\emph{vision})           & 65,536  & 0.194 &  0.516    & 87.52\\
    BL+PMDN (\emph{vision} + \emph{touch})   & 131,072 & 0.194 &  0.454 & 87.94 \\
    BL+RegBN (\emph{vision})  & 0         & 0.052 & 0.267 & 90.08\\
    BL+RegBN (\emph{vision} + \emph{touch})  & 0 & {0.052} & {0.231} & {91.54}\\
    \bottomrule
  \end{tabular}
\end{table}

\begin{table}
  \caption{Accuracy of classification results on the synthetic dataset in the presence of a confounder. Comparison of the reference to an approach without (W/O) normalization and several normalization techniques. Due to randomness, we reported the mean and std of results over 100 runs.}
  \label{table:appendix-synth}
  \centering
  \begin{tabular}{@{\extracolsep{4pt}}l cc cccc@{}}
    \toprule
    Normalization   &\multicolumn{2}{c}{Norm. params. ($\#$)}   & \multicolumn{2}{c}{Experiment I} & \multicolumn{2}{c}{Experiment II}  \\
    \cmidrule{2-3} \cmidrule{4-5}  \cmidrule{6-7} 
    Method &  MLP & CNN & MLP & CNN & MLP & CN \\
    \midrule
    \emph{reference} &      - &   -      & \emph{87.5}  & \emph{87.5}  & \emph{75.0} & \emph{75.0}\\
    W/O normalization  &  0  &  0  & 96.2±0.4 & 96.3±0.3 & 86.7±0.6 & 85.4±0.7\\
    BN         & 256 & 112 & 96.4±0.4 & 94.9±0.2 & 83.1±0.8 & 85.2±0.3\\
    GN         & 256 & 112  & 96.3±0.3 & 95.7±0.3 & 84.6±0.6 & 86.7±0.4\\
    LN         & 256 &  112 & 96.1±0.3 & 95.8±0.2 & 85.1±0.5& 84.7±0.3\\
    IN         & 256 & 112  & 96.4±0.4 & 95.9±0.2 & 84.6±0.4 & 83.2±0.9 \\
    MDN   &   0   &   0       & 91.9±0.7 & 89.7±0.7 & 79.1±0.6  & 82.2±0.7\\
    PMDN  &  2048 &   201,728 & 93.4±0.6 & 92.7±0.5 & 80.8±0.7  & 84.3±0.8\\
    RegBN &   0   &   0       & 87.3±0.9 & 88.2±0.8 & 76.2±0.8  & 76.8±0.9\\
    \bottomrule
  \end{tabular}
\end{table}

\subsection{Synthetic dataset} \label{appendix:synthResults} 
Table~\ref{table:appendix-synth} reports the accuracy results on the synthetic dataset for different batch normalization methods. 
The results obtained in this section are consistent with those reported in healthcare experiments. Just like in healthcare data analysis, where the multimodal data's confounding effects and unique characteristics pose challenges for conventional normalization methods, the findings in this section indicate that specialized normalization techniques, such as those mentioned, are necessary to effectively handle multimodal data. 
It is worth noting that metadata frequently incorporates confounding variables and noisy data that may be unknown or difficult to measure. In a complicated scenario, the metadata can not only affect the training procedure
but also correlates with the predicted label~\cite{lu2021metadata}. 
Therefore, it is crucial for a network to differentiate and handle such values effectively. 
In the case of employing the MDN and PMDN approaches, it is obligatory to explicitly declare both the metadata and the corresponding labels.
However, unlike MDN and PMDN, there is no requirement to specify the metadata and labels when applying RegBN to multimodal models.

\section{Ablation study}\label{appendix:abalation}
Here we investigate the influence of RegBN's parameters on the training and test results.

\begin{figure*}
\centering
    \subcaptionbox{Validation accuracy with Experiment I\label{fig:appendix-lambdafixedadapAccI}}{\includegraphics[width=0.65\textwidth]{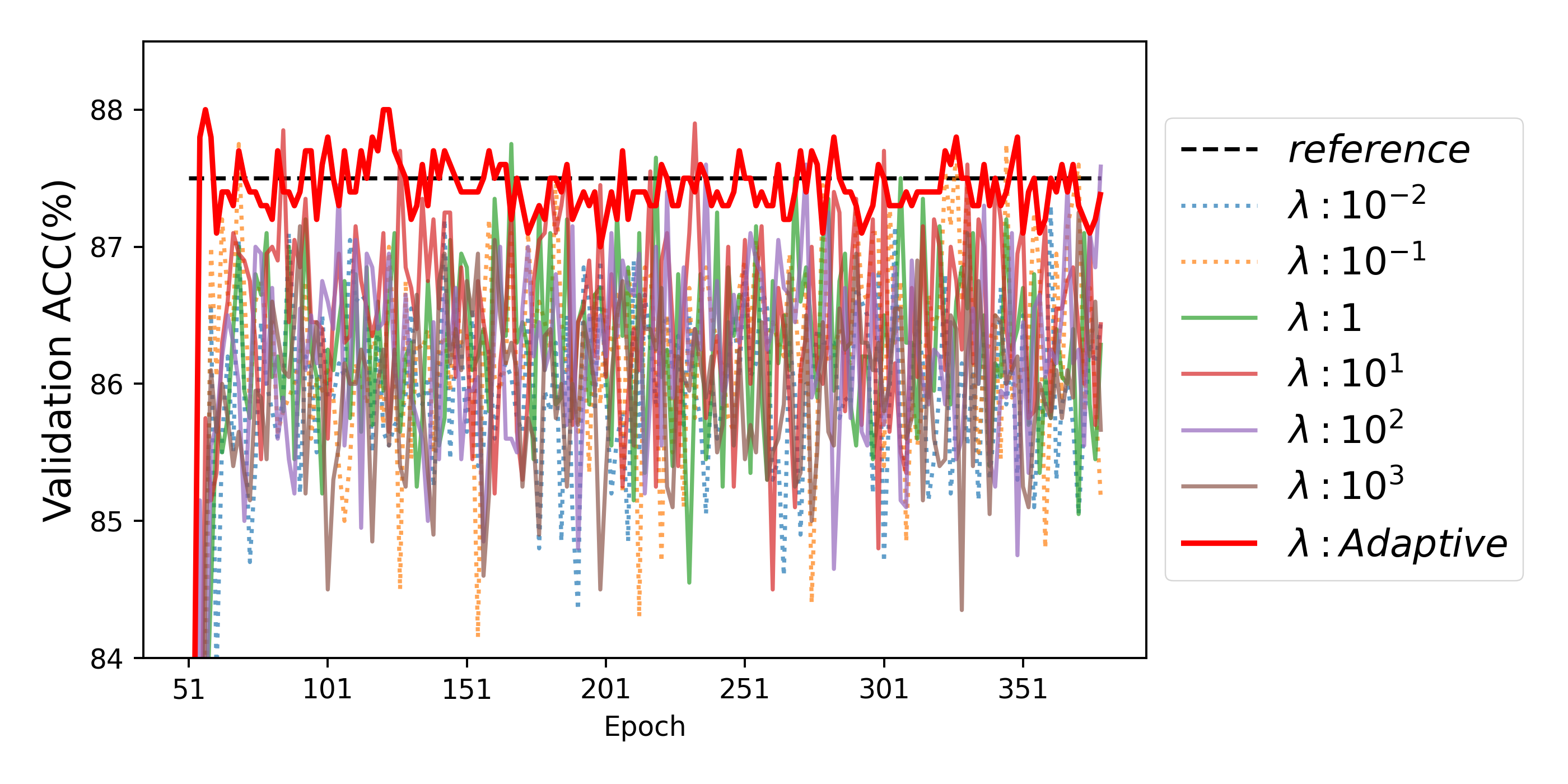}}\hspace{0em}
    \subcaptionbox{The training loss with Experiment I\label{fig:appendix-lambdafixedadapLossI}}{\includegraphics[width=0.65\textwidth]{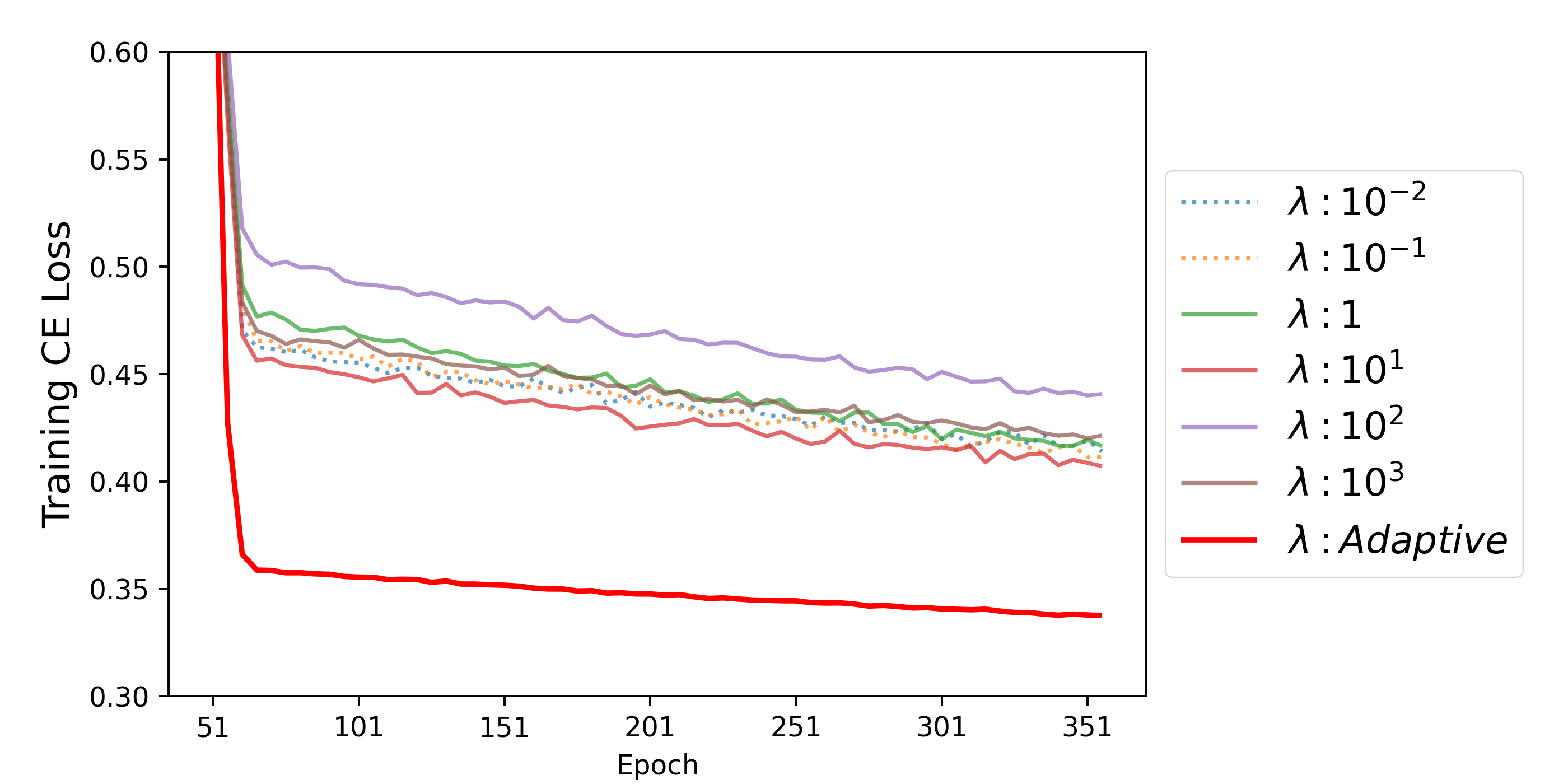}}\hspace{0em}
    \subcaptionbox{Validation accuracy with Experiment II\label{fig:appendix-lambdafixedadapAccII}}{\includegraphics[width=0.65\textwidth]{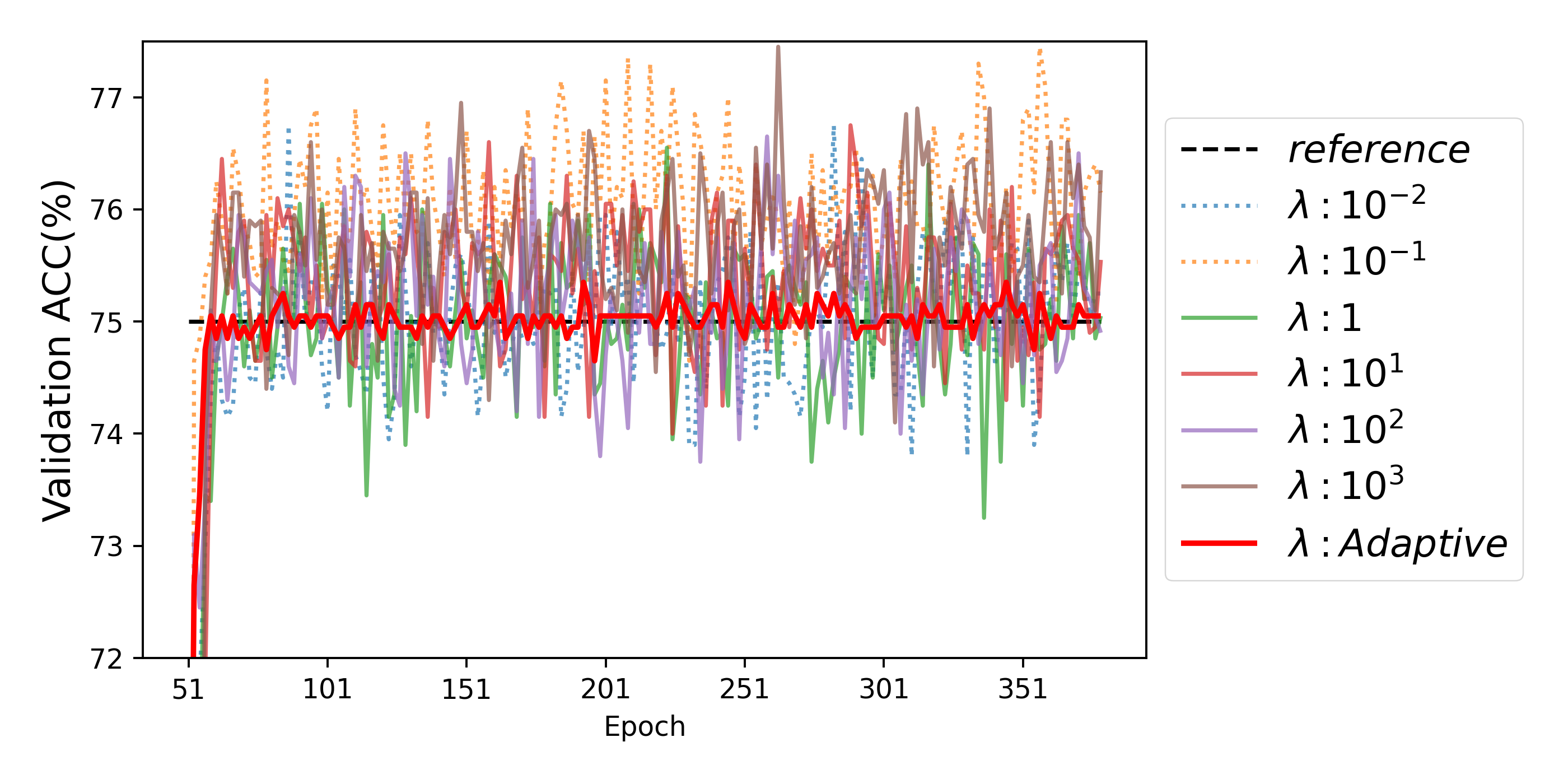}}\hspace{0em}
    \subcaptionbox{The training loss with Experiment II\label{fig:appendix-lambdafixedadapLossII}}{\includegraphics[width=0.65\textwidth]{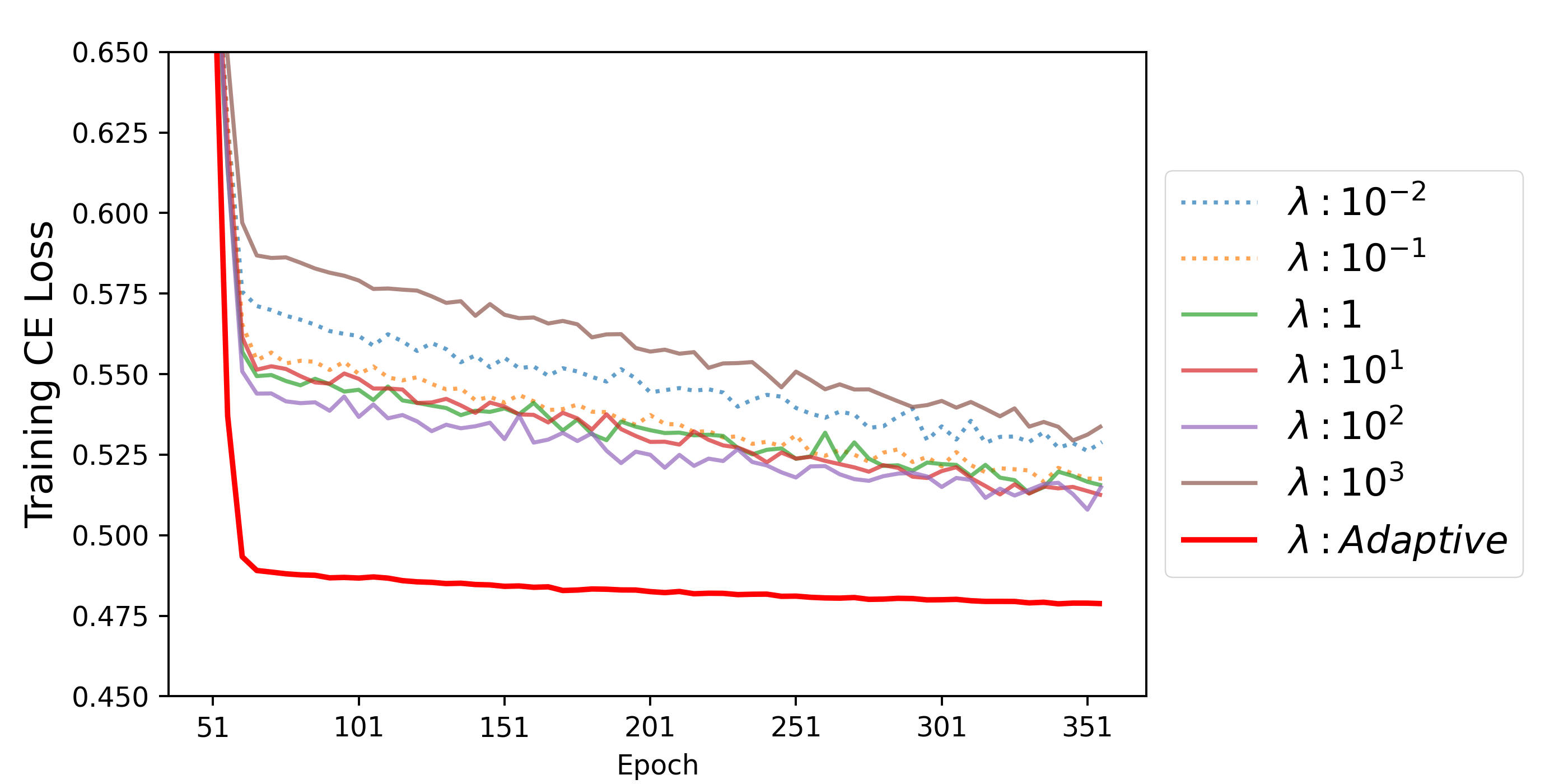}}\hspace{0em}
\caption{Impact of fixed and adaptively-estimated $\lambda_+$ values in RegBN on the synthetic dataset.}
\label{fig:appendix-ablation-lambdaFixAda}
\end{figure*}

\textbf{Comparing adaptive method for estimating $\lambda_+$ with fixed $\lambda_+$ values}: 
In Section~\ref{sec:avoidextrema}, we introduced a recursive method for estimating $\lambda_+$ values in each mini-batch to prevent falling into local minima. This section examines the effects of using fixed $\lambda_+$ values versus adaptively-estimated ones. To this end, we selected the synthetic experiment, wherein the amount of confounding effect is known and measurable. 
Figure~\ref{fig:appendix-ablation-lambdaFixAda} illustrates the results obtained for both Experiments I and II. It is evident that when using fixed $\lambda_+$ values, the confounding effect still persists in the results. However, the proposed adaptive technique successfully tracks and removes the confounding effect, leading to a smoother training loss and convergence of the multimodal data toward its reference value. 
Furthermore, the validation accuracy obtained with the adaptive method demonstrates a narrower range of perturbation in accuracy, indicating the effectiveness of this approach in removing the confounding effect. The adaptive technique proves to be efficient in addressing the challenges posed by confounding factors and achieving more reliable and stable results in the presence of such effects.

\begin{figure*}
\centering
    \subcaptionbox{Range of $\lambda_+$ values in Experiment I\label{fig:appendix-lambda_quad_expI}}{\includegraphics[width=0.98\textwidth]{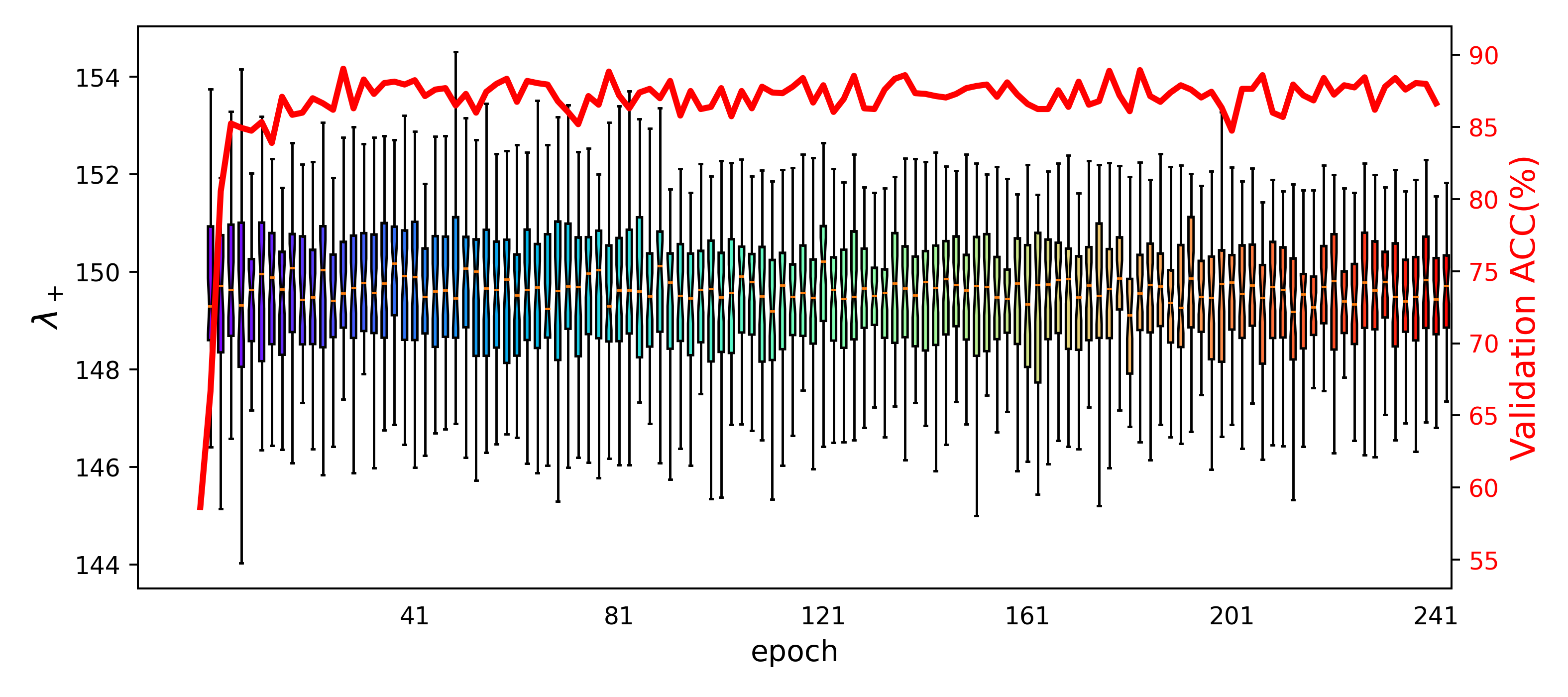}}\hspace{0em}
    \subcaptionbox{Range of $\lambda_+$ values in Experiment II\label{fig:appendix-lambda_quad_expII}}{\includegraphics[width=0.98\textwidth]{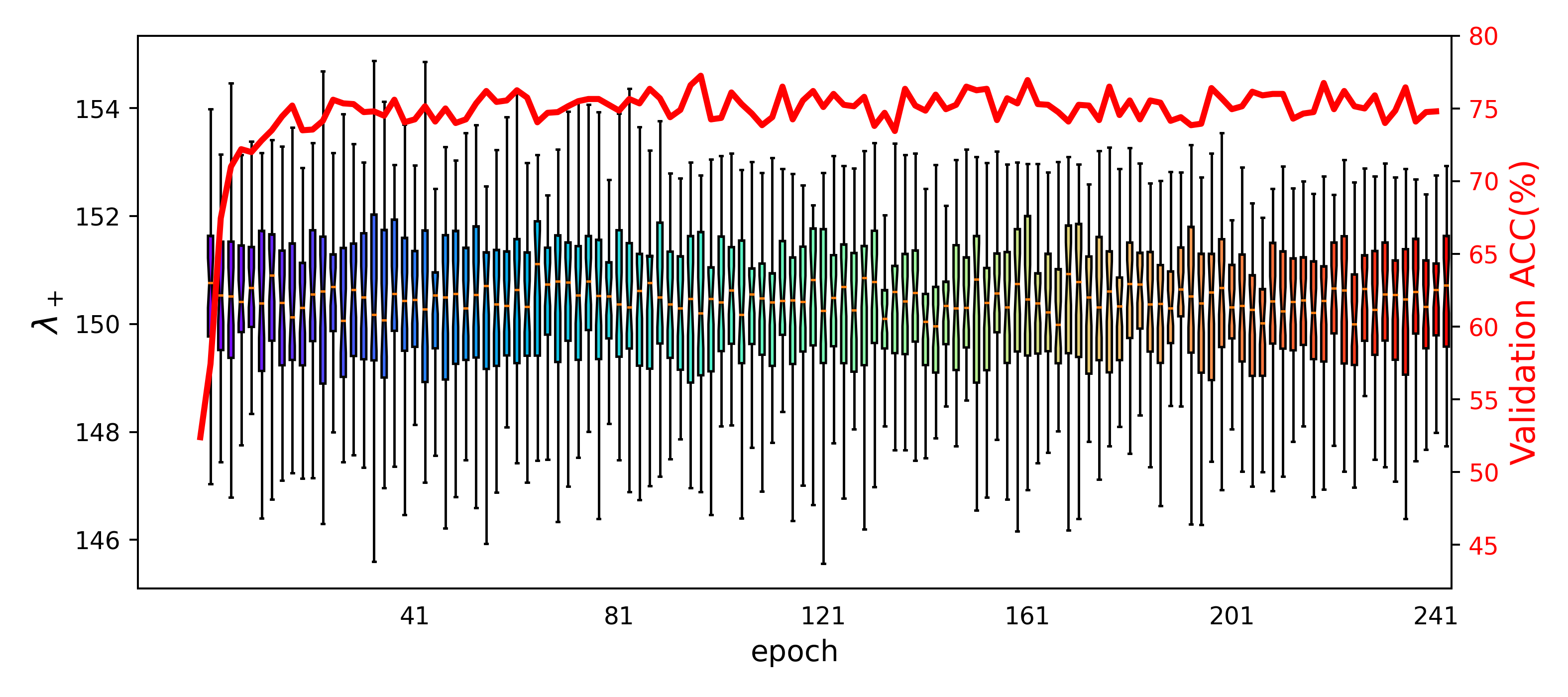}}\hspace{0em}
\caption{Evolution of the distribution and variability of the estimated $\lambda_+$ values across epochs in the synthetic dataset using boxplots. 
The estimated $\lambda_+$ values exhibit a progressively decreasing range as the training progresses. 
}
\label{fig:appendix-ablation-lambda-quad}
\end{figure*}

\textbf{Varaiation of $\lambda_+$ values over mini-batches}:
As discussed in the aforementioned experiment and also in Section~\ref{sec:proposed}, the projection matrix in RegBN is estimated in every mini-batch (due to the adaptive estimation of $\lambda_+$) and updated recursively. 
To visualize the evolution of the estimated $\lambda_+$ values throughout the training process, Figure~\ref{fig:appendix-ablation-lambda-quad} displays a box plot representing the distribution of computed $\lambda_+$ values over epochs in the synthetic dataset. 
In both Experiment I and Experiment II, the box and whisker plots exhibit a noticeable reduction in length as the multimodal model is trained further. Larger whiskers and wider box plots are observed during the initial epochs, indicating a wider spread of $\lambda_+$ values. However, as the training progresses, the $\lambda_+$ values converge across the epochs, resulting in shorter whiskers and box plots. The $\lambda_+$ values become increasingly consistent and less variable as long as the multimodal model converges.

\begin{figure*}
\centering
    \includegraphics[width=0.8\textwidth]{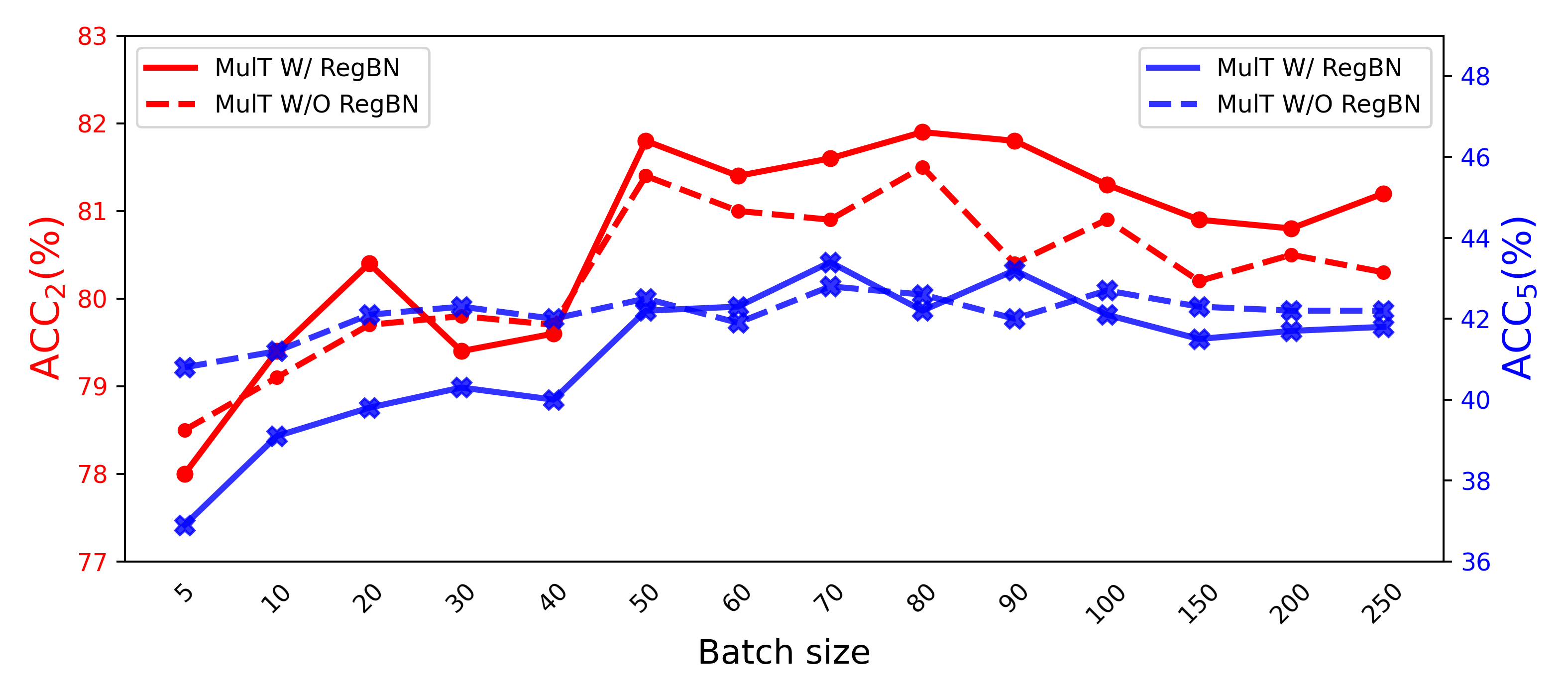}
\caption{Effect of batch size on the accuracy of MulT~\cite{tsai2019multimodal} with/without RegBN in the CMU-MOSI dataset. 
}
\label{fig:appendix-ablation-bs}
\end{figure*}

\textbf{Batch size}: In this study, a fixed batch size of 50 was utilized for all experiments with RegBN. Figure~\ref{fig:appendix-ablation-bs} presents the influence of batch size on the binary and 5-class accuracy obtained with MulT~\cite{tsai2019multimodal} with/without RegBN in the CMU-MOSI dataset. 
The figure indicates that batch sizes of 50 and above yield more favorable outcomes when employing RegBN as a normalization method. 
On the other hand, when using smaller batch sizes, the desired outcomes may not be achieved due to two underlying reasons. Firstly, for RegBN to accurately estimate and optimize its projection matrix, a sufficient number of observations are required, which is not feasible with small batch sizes. Secondly, there is an interaction between the convergence of the multimodal model and the convergence of RegBN's projection matrix, and they influence each other. 
The MulT technique does not yield good results for batch sizes of 40 or lower, which directly impacts the performance of RegBN. Generally, to ensure reliable estimation of the projection matrix and subsequent normalization, a batch size of 50 or higher is recommended for RegBN, providing adequate observations. 
In contrast to MDN, which is susceptible to the negative effects of small batch sizes, RegBN demonstrates effective functionality even with lower batch size values.

\textbf{Impact of L-BFGS parameters on RegBN}: In each mini-batch, RegBN seeks the best estimation for $\lambda_+$ via L-BFGS. The learning rate and the maximum number of iterations per optimization step are the important predefined parameters of the L-BFGS algorithm. 
Table~\ref{table:appendix-BFGS} reports the binary accuracy (Acc$_2$) and 5-class accuracy (Acc$_5$) of the MulT method with RegBN on the CMU-MOSI dataset. 
It is recommended to use a learning rate within the interval of [0.5, 1.0] and configure the maximum number of iterations per step to fall within the range of [20, 30].

\begin{table}[t]
  \caption{Effect of learning rates and maximal steps on the accuracy of MulT~\cite{tsai2019multimodal} with RegBN in the CMU-MOSI dataset.}
  \label{table:appendix-BFGS}
  \centering
  \begin{tabular}{l cc | l cc }
    \toprule
    Learning rate  &Acc$_2$$\uparrow$ & Acc$_5$$\uparrow$ & Max. Iterations (\#) &Acc$_2$$\uparrow$ & Acc$_5$$\uparrow$ \\
    (\#Max. Iterations=25) &&& (Learning rate=1.0)
    \\
    \midrule
    0.01  & 80.7 & 43.2   & 5  & 80.9 & 41.3\\
    0.1   & 80.6 & 42.9   & 10 & 81.1 & 42.1\\
    0.5   & 81.0 & 42.7   & 15 & 81.2 & 41.7\\
    1.0   & 81.8 & 42.3   & 20 & 81.6 & 42.5\\
    1.5   & 80.8 & 41.4   & 25 & 81.8 & 42.3\\
    2.0   & 80.2 & 41.2   & 30 & 81.9 & 42.3\\
    5.0   & 80.8 & 41.3   & 40 & 81.8 & 42.1\\
    10.0  & 80.5 & 42.7   & 50 & 81.9 & 42.0\\
    \bottomrule
  \end{tabular}
\end{table}

\pagebreak
\section{Questions and Answers}\label{appendix:QA}

\begin{enumerate}[
    leftmargin=*,
    label={Q\arabic*\ },
    ref={Q\arabic*}]
  \item {How do the concepts of multimodal normalization, consistency, and complementarity interconnect?} 
  
  \textit{Answer}: \textbf{Complementarity} refers to unique information from different modalities, and their combination enriches overall multimodal data interpretation. \textbf{Consistency} refers to the degree to which information, patterns, or features align and agree across different modalities' data. In other words, consistency assesses how well the information conveyed by such different modalities corresponds or converges towards a shared understanding. For instance, consistency in the context of vision-text is the textual descriptions accurately represent the content depicted in the images, and vice versa. In the best scenario, information from a modality must reinforce and complement the information from other modalities, leading to a coherent and unified interpretation. \textbf{Multimodal nomalisation} like RegBN, which is introduced in this study, aims at making the different modalities' data independent by removing confoundings. As demonstrated by the quantitative and qualitative experimental results, ensuring independence can improve the reliability of analyses and predictions by leveraging the synergies between different types of information while minimizing confounding impacts between modalities.
  
  \item {Could RegBN improve the \emph{modality imbalance} problem in multimodal databases?} 
  
  \textit{Answer}: Modality imbalance refers to an uneven distribution of performance or representation among different data modalities like image, audio, etc in multimodal learning. Though the whole multimodal network performance exceeds any single modality, each modality performs significantly below its optimal level. RegBN shows that harnessing modality independence is an efficient means to synergize diverse information types.

  \item {Is RegBN a fusion model?} 
  
  \textit{Answer}: No, RegBN functions by normalizing input X in relation to input Y, resulting in a normalized X with the same dimensions as the original input X. In fusion, inputs X and Y are combined to generate one or more outputs with distinct content and dimensions. RegBN can be used as a normalization method within the structure of any fusion or neural network.

  \item {Is it feasible to implement RegBN multiple times within a neural network?} 
  
  \textit{Answer}: Yes, RegBN, like other normalization techniques, can be used multiple times in neural networks.  
  RegBN acts as an independence-promoting layer, so utilizing it multiple times in a row does not substantially alter the feature maps. 
  For a pair of modalities, as outlined in Section~\ref{appendix:fusion}, it is advised to employ RegBN once as a multimodal normalizer within various fusion paradigms. 
  It is important to highlight that in the context of layer fusion (Figure~\ref{fig:appendix-fusionmid}), where RegBN is employed multiple times, the input feature maps at each instance differ from one another. RegBN can be employed as long as its inputs are not mutually independent.

\end{enumerate}

\end{document}